\documentclass[journal,twocolumn,a4paper,10pt]{IEEEtran}

\usepackage{graphicx}
\usepackage{amsmath}
\usepackage{amssymb}
\usepackage{amsfonts}

\newcommand{\dnorm}[1]{{\bigl\|{#1}\bigr\|}}

\usepackage[latin1]{inputenc}
\usepackage{subfig}

\usepackage{theorem}
\theoremstyle{break}{\theorembodyfont{\itshape} }{\theorembodyfont{\itshape} }

\title{Modelling, Measuring and Compensating Color Weak Vision}

\author{Satoshi Oshima, Rica Mochizuki, Reiner Lenz, Jinhui Chao     
\thanks{S. Oshima is with NTT Advanced Technology, Tokyo, Japan.}
\thanks{R. Mochizuki is with NTT Solution Lab, Tokyo, Japan.}
\thanks{R. Lenz is with Dept. Science and Technology and the Dept. Electrical Engineering Link\"oping University, SE-60174 Norrk\"oping, Sweden.}
\thanks{J. Chao is with Dept. of Information and System Engineering, Chuo University, Tokyo, Japan.}}

\begin{document}

\maketitle

\begin{abstract}
We use methods from Riemann geometry to investigate transformations between the color spaces of color-normal and color weak observers. The two main applications are the simulation of the perception of a color weak observer for a color normal observer and the compensation of color images in a way that a color weak observer has approximately the same perception as a color normal observer. The metrics in the color spaces of interest are characterized with the help of ellipsoids defined by the just-noticable-differences between color which are measured with the help of color-matching experiments. The constructed mappings are isometries of Riemann spaces that preserve the perceived color-differences for both observers. Among the two approaches to build such an isometry, we introduce normal coordinates in Riemann spaces as a tool to construct a global color-weak compensation map. Compared to previously used methods this method is free from approximation errors due to local linearizations and it avoids the problem of shifting locations of the origin of the local coordinate system. We analyse the variations of the Riemann metrics for different observers obtained from new color matching experiments and describe three variations of the basic method. The performance of the methods is evaluated with the help of semantic differential (SD) tests. 
\end{abstract}
\section{Introduction}
As many as eight percent of the male population and 0.4~percent of women are color blind and even more suffer some kind of color vision defects in various degrees. We will refer to these less severe color vision defects as color-weakness. Apart from these cases of deficient color vision it is also obvious that there are significant variations of the color vision properties of persons with normal color vision. In critical applications, such as the design traffic signals or natural disaster alarm, one has to make sure that even color blind persons can easily recognize these symbols. Very often this is achieved by enhancing the color contrast between the symbol and the background. Obviously this is not possible for natural images.

Based on this description of color-vision related problems one could formulate the ultimate goal of an adaptive color presentation between two types of observers as follows: Given a stimulus $S$, generate a modified stimulus $S'$ such that the perception of observers of type one viewing $S$ is the same as the perception of observers type two viewing $S'$. Two examples would be methods to simulate the perception of color weak or color blind observers and methods to compensate the effects of color weak deficiencies. 

In this generality the problem cannot be solved, simply because it is impossible to compare color perceptions of different observers objectively. In fact, even on the sensor-related, low-level we cannot measure quantitative properties of human perception with objective and non-invasive methods and it is therefore impossible to characterize the color-weakness of an observer exactly. 

One approach that avoids these problems is the usage of color matching based methods which characterizes the properties of the color perception of an observer by measuring local color differences (see~\cite{goldstein,Koenderink}). In this paper we investigate methods for color-weakness compensation based on the equivalence between all perceptional differences among the color distribution in the color spaces of both the color normal and the color-weak observer. In terms of local geometry, this means that we match color discrimination thresholds of the color-weak and the color-normal observer. These small color differences or discrimination thresholds (typical examples are the MacAdam ellipsoids) are one of the fundamental color vision characteristics and among the few observables in color perception which can be used to characterize the perceptual characteristics of individual observers. Such a discrimination ellipsoid describes the just-noticeable difference (jnd) between the color at the center of the ellipsoid and other colors in its neighborhood. The applied methods were developed in the framework of Riemann geometry. In this framework these ellipsoids characterize the local geometry around center points, and the color space becomes a Riemann space with the thresholds defining the Riemann metric tensor. A map between color spaces that preserves the local discrimination thresholds everywhere is a local isometry. A global isometry or a compensation map can be obtained by integrating these local mappings using tools from Riemann geometry. Details of a method based on this strategy can be found in~\cite{CGIV2008-colorweak,HCI2011,CCIW2011}. For special types of color weakness, 1-D compensation methods in closed forms and their fast implementation are described in~\cite{CCIW2011} and evaluations of this approach are reported in~\cite{Chen-1} and~\cite{Chen-2}. These mappings use techniques from linear algebra and they are therefore easy to implement. However, for many color-weak observers these methods are not general enough and simultaneous compensations in more than one dimension are necessary. There is thus a need for 2D and 3D methods similar to the ones based on these local affine maps. Unfortunately straight forward generalizations have a limited performance due to the approximation errors originating in the local linearization. An additional problem is the estimation of the corresponding positions of the origins of the local neighborhoods. 

In this paper we present a compensation without local linearization based on the global Riemann geometry of 3D color spaces that avoids these problems. We use the fact that if a mapping is a local isometry for all points in the color space then it is also a global isometry. We will introduce a procedure to construct a global isometry by using Riemann normal coordinates which consist of geodesics in (2D and 3D) color spaces. 

The construction of the isometry requires the knowledge of the metric tensor. We acquire the necessary data by measuring color discrimination threshold data for both color-normal and color-weak observers. In~\cite{CCIW2009} discrimination threshold data was measured on the chromaticity plane of the CIEXYZ space and a 2D compensation was implemented. The experiments showed that more accurate threshold data were required and that chromaticity information alone was, in general, not sufficient to produce satisfying compensation results. This showed that a simultaneous compensation of both lightness and chromaticity in 3D color spaces is needed. Here we describe a new algorithm for fast and stable measurement of the 3D threshold data. As a result we obtain a new set of discrimination threshold measurements for both color-weak and normal observers in CIELUV space of reasonably high resolution. We describe this new dataset and we then discuss the implementation of the color-weak compensation, based on these new measurements, regarding the cost and performance. We consider (in increasing order of complexity) three algorithms: 2D, 2D+1D and 3D compensation maps. This results in global isometries or color-weak maps, that simulate the color-weak vision for a color normal observer. Its inverse can be used as compensation map that transforms a given image as so that the color-weak observer can experience a similar color perception as the corresponding color-normal observer. We apply the method to natural images and evaluate its performance using semantic differential (SD) tests~\cite{SD_method}.

\section{Riemann geometry of color spaces}
An $n$-dimensional Riemann space is a space that locally looks like a Euclidean space $\mathbb{R}^n$ with a metric which is defined by an $n$ by $n$ matrix $G$ such that the squared length of a vector $x$ is defined as $x^TGx$ 
where~$(\cdot)^{T}$ denotes vector transposition.~(\cite{Color-Science,DoCarmo}). 
In a Riemann space, the local geometry is described by the metric tensor~$G(x)$ for every point~$x$ in the space where $G(x)$ is a positive-definite symmetric matrix smoothly varying along $x$ defining the local distance around~$x$. The matrix~$G(x)$ defines a quadratic form on the tangent space at~$x$ where the squared length of a tangent vector~$dx$ is given by
\begin{equation}
\dnorm{dx}^2 = dx^{T}G(x)dx.
\end{equation}

Actually, color vision was one of the application areas mentioned by Riemann when he introduced his new geometrical construction and Wright's intervals and MacAdam ellipses and ellipsoids are among the first measurements of the metric tensor in 1D, 2D and 3D spaces (see~\cite{Color-Science},\cite{macadam} for a historical overview).

In color science the local metric is estimated with the help of color matching experiments. In such an experiment an observer compares two colors, a fixed test color and a second color. At the beginning of the experiment the two colors are identical. The observer can control the second color and move it a certain direction in a color space until he/she can see a difference between the test color and the modified color. This color difference is known as the just-noticeable-difference (jnd) or  a color discrimination threshold. A series of such matching experiments, where the secondary color is moved in different directions in color space, can be used to construct the matrix~$G(x)$. Details of this construction and a new version will be described later in the experimental part of the paper. 

The distance~$d(x_1,x_2)$ between two points $x_1, x_2$ in a Riemann space is defined as the length of the shortest path $\gamma_{12}$ connecting the two points. These paths are the geodesics and large color differences in color spaces can be defined is this distance on the Riemann manifold:
\begin{equation}
d(x_1,x_2)=\int _{\gamma_{12}}\dnorm{dx} =\int _{\gamma_{12}}\sqrt{dx^T G(x)dx}.
\label{eq:geodesic}
\end{equation}

We model the color perception of an individual observer by deriving the metric of the color space from his/her threshold data and subjective color differences are then measured as distances in the obtained color space as a Riemann space. By comparing and preserving the distance between points in the color spaces of two observers we can compare and preserve subjective properties of the color vision of these observers. A map that preserves distances between two Riemann spaces is known in geometry as an isometry. A local isometry is defined as a map between two spaces~$C_1$ and~$C_2$:
\[ f: C_1 \longrightarrow C_2: x\longmapsto y=f(x) \]
that preserves local distances in the neighborhood of all points. This implies that $f$~maps space $C_1$ with Riemann metric $G_1(x)$ to space $C_2$ with metric $G_2(y)$ such that for every $x$~in $C_1$
\begin{equation}
G_1(x)=(D_f)^TG_2(y)D_f ,
\label{eq_local-isometry}
\end{equation}
where $D_f$  is the Jacobian of $f$. Local isometries are functions that map the threshold ellipsoid at every $x$ onto the corresponding threshold ellipsoid at $y=f(x)$.
A global isometry is a map~$f$ such that the distances $d(x_1, x_2)$ for every pair of $x_1, x_2$ in $C_1$ is the same as $d(y_1, y_2)$ for $y_1=f(x_1), y_2=f(x_2)$ in~$C_2$. It can be shown that a local isometry  also gives a global isometry and vice versa.
\section{Compensation of color weak vision with isometries}

So far we described how color matching experiments provide input data that can be used to construct a Riemann manifold model that characterizes the properties of the low-level part of the color vision system of an observer. In this framework isometries can be constructed that preserve the geometric structure of the color spaces involved. The assumption is that with the help of such a mapping we  can generate a visual input that allows us to convey the color perception of one observer to another observer. As a result it should be possible to share the perceptional experience between observers (at least on this low-level of color perception). Constructing color-difference-preserving maps~$f$ between observers is then equivalent to the construction
 of isometries between Riemann manifolds.

As mentioned before, we are mainly interested in the relation between the color perception of color-weak and color-normal observers.  
We define a color-weak map $w$ as an isometry from the color space $C_w$ of the color-weak observer to the space $C_n$ of  the color-normal observer as
\[
w : C_w \longrightarrow  C_n : y = w(x)
\]
This map shows to the color-normal observers what is actually seen by the color-weak observer and provides thus a color-weak simulation map, which simulates the color-weak perception from a color-normal's point of view. The inverse map of the color-weak map $w$ is also an isometry, now from the color
 space of the color-normal observers to the color space of a color-weak observer
\[
w^{-1} : C_n \longrightarrow  C_w : x = w^{-1}(y)
\]
which shows to the color-weak observer what the color-normal observer actually sees, it is thus a compensation map.

There are two ways to construct isometries: one can either match discrimination threshold ellipsoids by local affine maps or one can build the global isometry between Riemann spaces directly. The method using local affine maps has the advantage to it is easier to understand and easier to implement using
 basic linear algebra. Descriptions of this approach can be found in~\cite{CGIV2008-colorweak} and~\cite{HCI2011}. There are two major problems with this approach: numerical errors due to approximation in linearization are significant and it is difficult to match the positions and the correspondences
 between the origins of the local neighborhoods in the two spaces. The second approach to compute the isometry directly is also difficult since the proof
 of the existence of a global isometry gives very few clues regarding the practical implementation. In the following we will introduce a method to build a
 global isometry which uses Riemann normal coordinates. First attempts to implement this method are reported in \cite{CGIV2004} and \cite{CGIV2008-Riemann}.

 We will first consider the special case where we make use of the Riemann normal coordinates in a Riemann space to obtain an isometry between the Riemann space and the Euclidean space. We call this an uniformization from the color space~$C$ to the Euclidean space~$U$ and denote it by 
$p$. Now, instead of constructing the isometry between color spaces~$C_i, i=1,2$ directly we will construct first uniformizations~$p_i$ to the common
 Euclidean space~$U$. Since the inverse of an isometry and composition of two isometries are also isometries, we obtain the isometry between $C_1$ and $C_2$
 as $f=p_2^{-1}\circ p_1$. The Riemann normal coordinates can be regarded as a generalization of the polar coordinates in the Euclidean space and we will construct the isometric image of the polar coordinates in the Euclidean space. 

\begin{figure}[htb]
 \begin{center}
  \includegraphics[width=0.9\columnwidth]{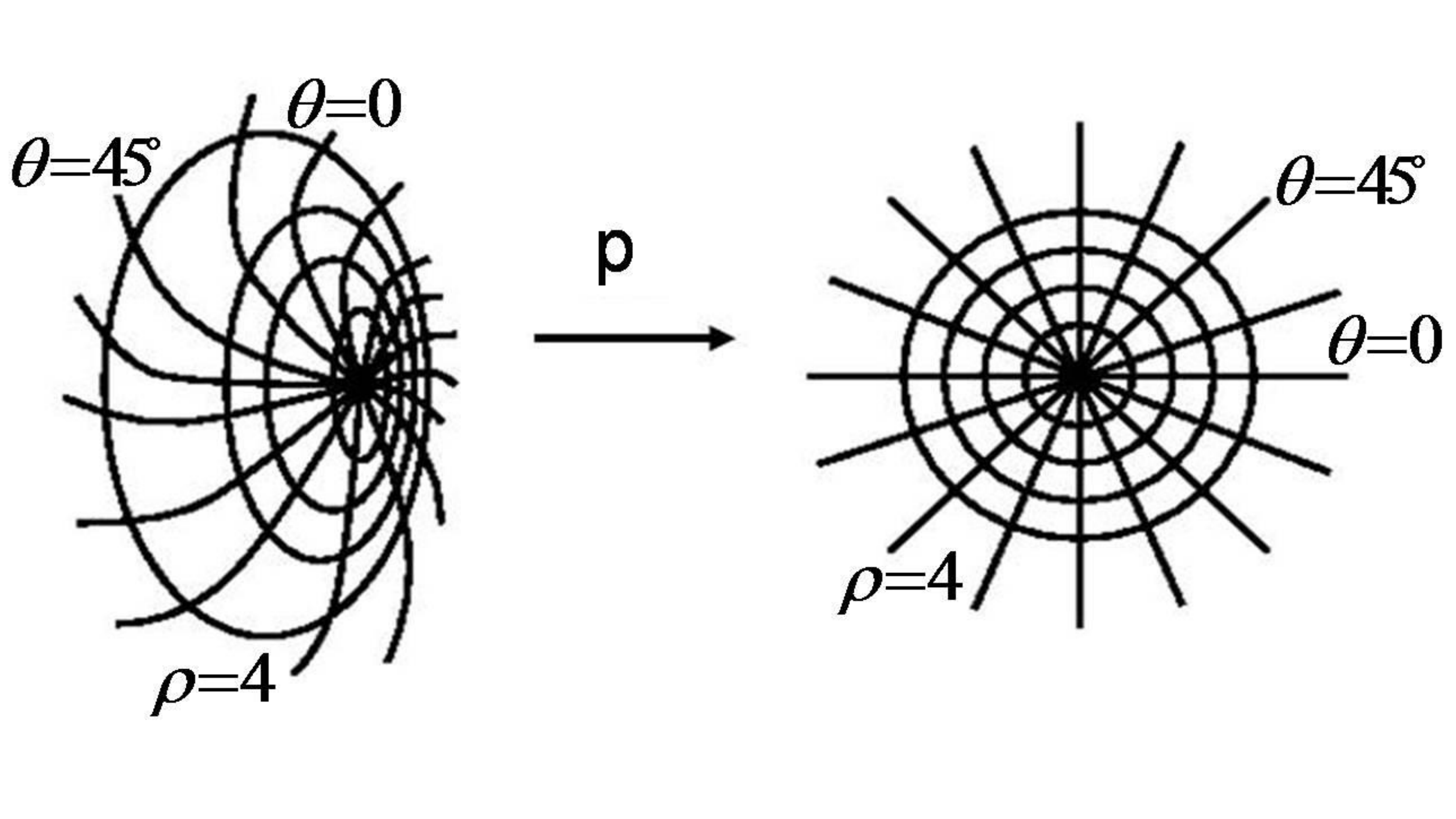}
  \caption{Uniformization of a color space using Riemann normal coordinates}
  \label{fig:ucs_isometry}
 \end{center}
\end{figure}
\begin{figure}[htb]
  \begin{center}
    \includegraphics[width=0.9\columnwidth]{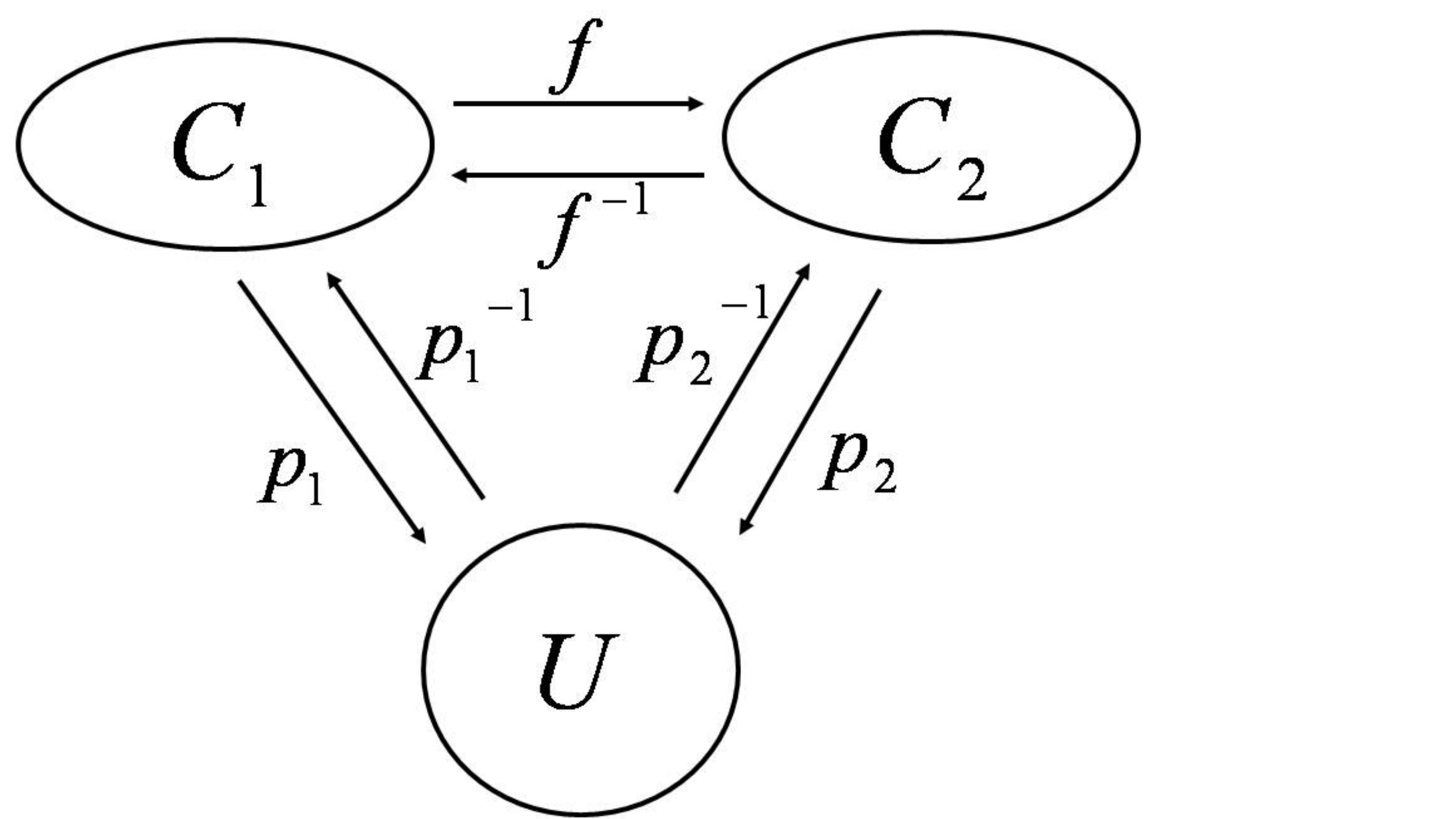}
  \caption{Construction of global isometry between two arbitrary spaces}
  \label{fig_global_isometry}
    \end{center}
\end{figure}
\section{Construction of isometries by Riemann normal coordinates}

In Eq.(\ref{eq:geodesic}) we introduced the distance between two colors as the length of the geodesic connecting the two points. A geodesic~$u$ in a Riemann space corresponds to a straight line in a Euclidean space. It can be shown~\cite{DoCarmo} that it is given by the solution of the differential equations (in the local coordinates~$u^i$)
\begin{equation}
\frac{d^{2}u^{i}}{ds}+\Gamma ^{i}_{jk}\frac{du^{j}}{ds}\frac{du^{k}}{ds}=0,
\end{equation}
where $\Gamma ^{i}_{jk}$ is the Christoffel symbol defined as: 
\begin{equation}
\Gamma ^{i}_{jk}=
\frac{1}{2}g^{i\alpha}\left(\frac{\partial g_{\alpha j}}{\partial u^{k}}+\frac{\partial g_{\alpha k}}{\partial u^{j}}-
\frac{g_{jk}}{\partial u^{\alpha }}\right).
\end{equation}
As usual the metric tensor and its inverse are denoted by 
\[G=(g_{ij}),G^{-1}=(g^{ij})\]
and the Einstein summation convention is used.

Such a system can be solved by choosing an origin and the initial conditions e.g. the speed at the origin. The origin can be chosen as certain reference point which does not change between color normal and color weak observer. Natural choices in color science are the points specified by the white points D65
 or D50.

The initial conditions for the geodesics eminating from the origin could include the requirement of a unit length speed vector (therefore the natural parameter or curve length is used for parameterization) and the condition that the geodesics should be separated in the equal angles (lengths of curves and the angles between them are measured using Riemann metric $G(x)$ at the origin). However, these conditions are not sufficient to uniquely specify the coordinates systems in two different spaces to have the same reference direction. In the 2D case we have to ensure that the directions of the zero degree geodesics are identical. In the 3D case, it is even more complicated since the Frenet formula for spatial curves states that one needs to specify both the direction of the tangent vector of the geodesic and a second vector either the acceleration or the torsion. In the following we use the invariant chroma in~\cite{Brettel} as the common reference directions for both color normal and color weak observers. In our case this is the lightness direction and e.g. the blue (S cone excitation) for the color-normal and color-weak observers of type $P$ and $D$  since for them the perception of blue is the same.

The geodesics can be used to build a coordinate system as follows: every point lies on one geodesic and we can thus describe the location of the point by first specifying on which geodesics it lies and then use its distance to the origin. Points with the same distance from the origin form a circle or sphere centered at the origin. In the current application such a circle/sphere defines an equi-chroma surface. In our implementation we calculate the Riemann normal
 coordinates of a given point by first finding the cell in the coordinates grid which contains the point, then the distance and angle of the point is determined e.g. by convex interpolation in the cell.

\section{Measurement of discrimination threshold data}

Classical psychophysical measurement methods are the method of limits, the method of adjustment and the method of constant stimuli (see~\cite{goldstein} for a description). Using the method of limits the experimenter presents stimuli stepwise in ascending or descending order and notes when the observer detects a change. In the method of adjustments the experimenter or the observer adjusts the strength of the stimuli in a continuous manner until the observer notices the change. In the method of constant stimuli the experimenter presents a number of stimuli in random order and notices the reactions of the observer. 

While the totally random measurement method is the most precise, it is also the most time consuming. On the other hand, randomization is necessary in order to avoid bias due to anticipation and adaptation or learning effects of observers. In our experiments we use a randomized adjustment method in pair-wise comparison experiments to determine the color discrimination thresholds. A session of color-matching starts with the display of the test color on the left and a comparison color on the right. The observer is asked to use either the mouse wheel or a keyboard to adjust the comparison color until it matches the test color as close as possible. An accepted match finishes the session. 

For a given test color the comparison color is randomly chosen on straight lines in 14~directions centered at the test color. The initial position of the comparison color and the increment or decrement along each direction are randomly chosen from certain ranges and/or a selection of discrete values. Thus the observer only decides if the next comparison color is more similar to the test color but the observer has no control over the magnitude of the change of the comparison color. The switching time to the new comparison color needs to be adjusted according to the movement speed of the mouse wheel or the rate of keyboard inputs. We also introduce random variations to this adjustment. We used a SyncMaster XL24 monitor by Samsung illuminated by Panasonic Hf~PREMIER~fluorescent lamps, a Munsell N5.5 background and a 10~degree viewing angle. The distance between the observer and the screen is 80cm, the size of the two test and the comparison color frames on the display is $14\times 14$ cm. In our experiments we measured color discrimination thresholds of a normal observer and a typical color-weak deuteranomalous observer (D..type, green-weak). We applied the randomized adjustment method in CIELUV sRGB space. We measured 77~sampling points, at 5~different lightness levels L=30, 40, 50, 60, 70. Data grids in the chromaticity planes of these levels contain 9, 13, 19, 20 and 16~points. An ellipsoid at a point is estimated from deviations in 14~directions from the origin.

\begin{description}
\item[Step 1]
In the pre-adaptation period of 5~minutes before the measurement, a neutral gray (Munsell N5:5) is shown on the whole display.
\item[Step 2]
The two~$14\times 14$cm frames with the test and comparison colors are shown on the display (10~degree vision field, 80cm distance, test color on the left, comparison color on the right). The test color is fixed during one session.
\item[Step 3]
The observer uses a mouse wheel or keyboard keys which change the CIELUV values of the comparison color based on the algorithm described above.
\item[Step 4]
Repeating Step 3, the observer tries to make the comparison color as similar as possible to the test color. 
\item[Step 5]
The above measurement will be repeated four times. The average value of these results will be used as discrimination threshold data.
\item[Step 6]
In a rest period of seven seconds between every four sessions a neutral gray N5:5 is shown on the whole display.
\item[Step 7]
The procedure from Step~2 to Step~6 are repeated for all test colors in the color space. This is repeated for at most 90~minutes after which the measuring session ends. The next session will continue no sooner than three days from the end of the current session.
\end{description}

\begin{figure}[hbt]
\begin{center}
\includegraphics[width=0.9\columnwidth]{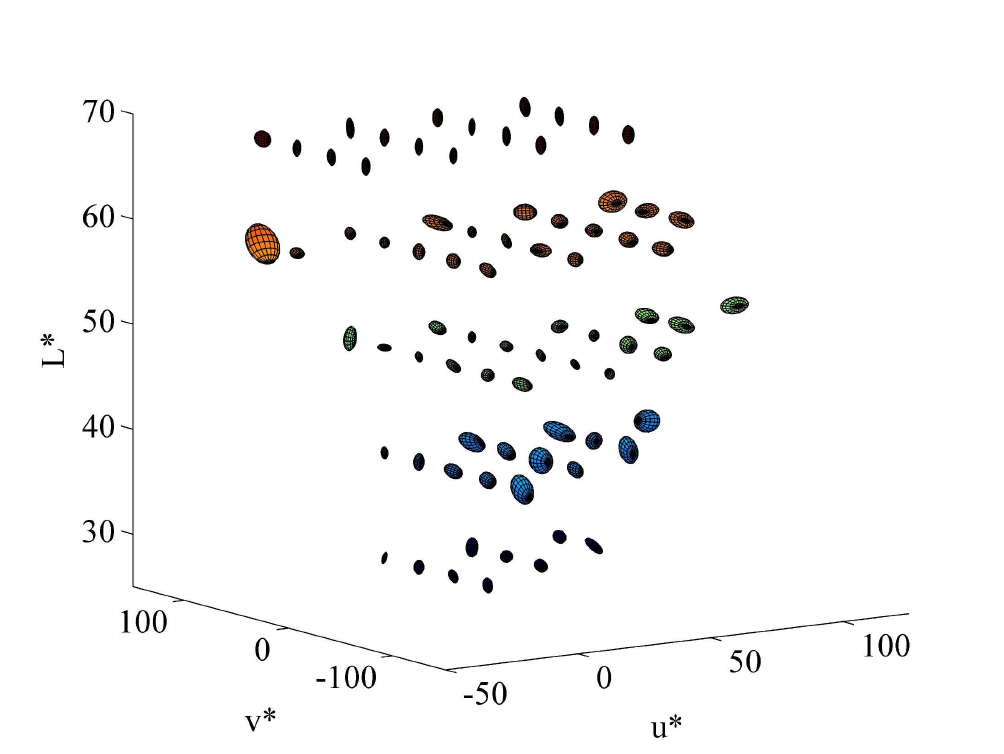}
\caption{Discrimination ellipsoids of a color-normal observer}
\label{fig:ellipsoids_normal}
\end{center}
\end{figure}

\begin{figure}[hbt]
\begin{center}
\includegraphics[width=0.9\columnwidth]{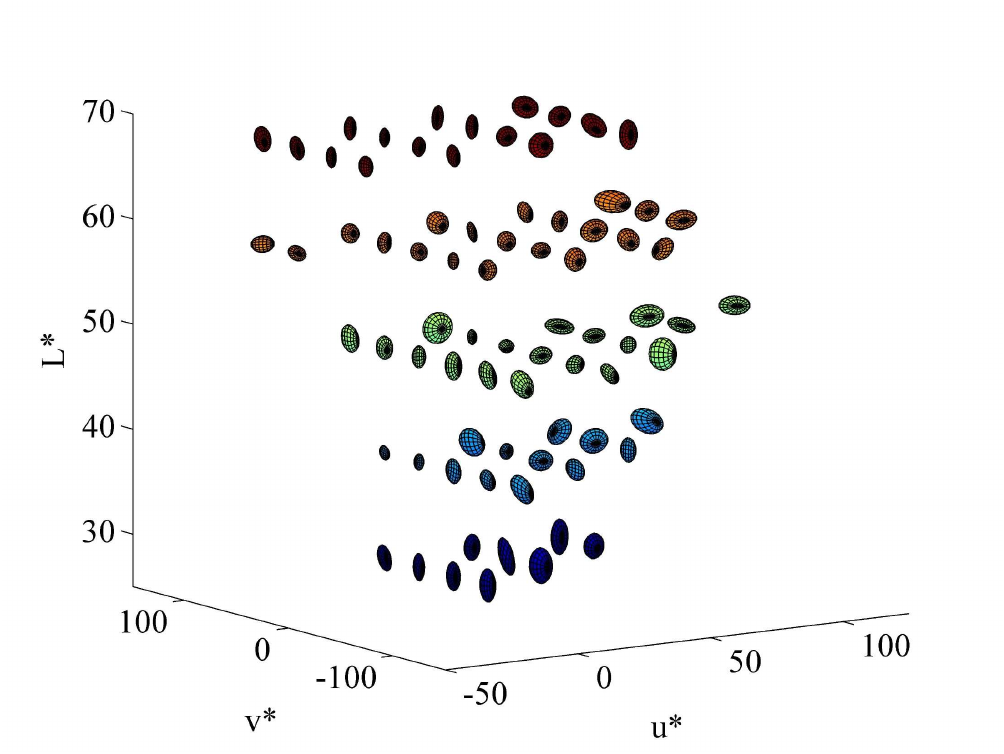}
\caption{Discrimination ellipsoids of a D-Type color-weak observer}
\label{fig:ellipsoids_weak_type_d}
\end{center}
\end{figure}

The measurements are repeated four times for each test color to obtain a statistically stable estimate. 
 The ellipsoids are then estimated from the observation data using the methods described in \cite{CGIV2008-colorweak}, \cite{ICCV2009} and~\cite{ICPR2010}. Discrimination threshold ellipsoids measured in 3D are
shown in Fig.~\ref{fig:ellipsoids_normal} for a color normal observer and Fig.~\ref{fig:ellipsoids_weak_type_d}. It can be seen that the discrimination threshold ellipsoids differ greatly in size and direction. Table~\ref{tab:ellipsoids_volume} shows the volume ratio between the ellipsoids of a color-normal observer and a color-weak observer.  It can be seen that the discrimination ellipsoids of a color-normal observer are on average 2.7 times larger than those of a color-weak observer. This explains the reduced color vision capabilities of the color-weak observer. One can also observe the differences of the shapes, sizes and orientations of the threshold ellipsoids between different lightness levels. 
 
\begin{table}[ht]
\begin{center}
\caption{Volume ratio of the ellipsoids between color-normal and color-weak observers}
\label{tab:ellipsoids_volume}
\begin{tabular}{c|c}
($L*$) & Average volume ratio \\\hline
$70$ & $3.1989$ \\
$60$ & $1.7918$ \\
$50$ & $3.2293$ \\
$40$ & $1.2089$ \\
$30$ & $4.8231$ \\
Overall & $2.6948$
\end{tabular}
\end{center}
\end{table}


\begin{figure}[htb]
\begin{tabular}{cc}
\begin{minipage}{0.5\textwidth}
\begin{center}
\includegraphics[width=0.9\columnwidth]{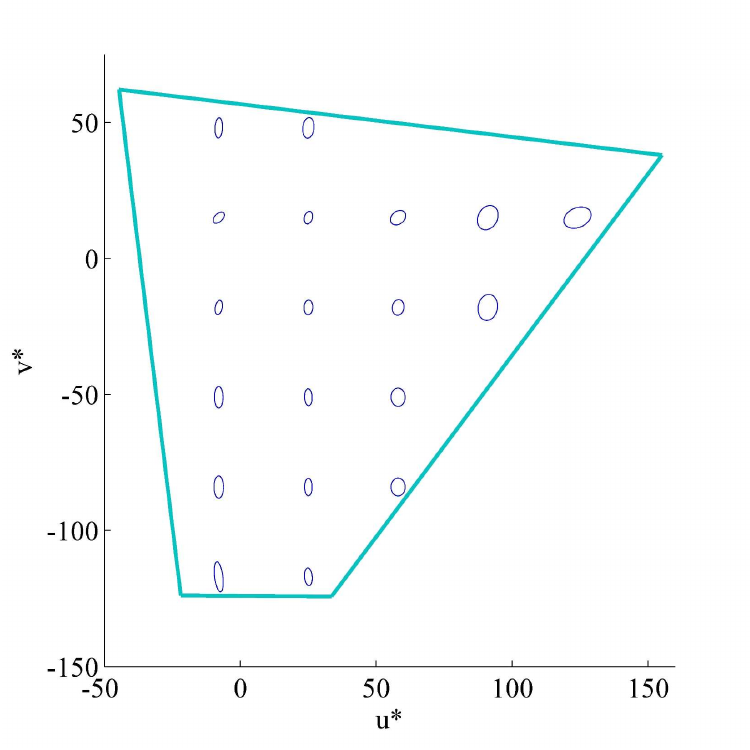}
\end{center}
\caption{Discrimination ellipses of a color-normal observer ($L^*=50$)}
\label{fig:type_c_ellipse_light3}
\end{minipage}\\
\begin{minipage}{0.5\textwidth}
\begin{center}
\includegraphics[width=0.9\columnwidth]{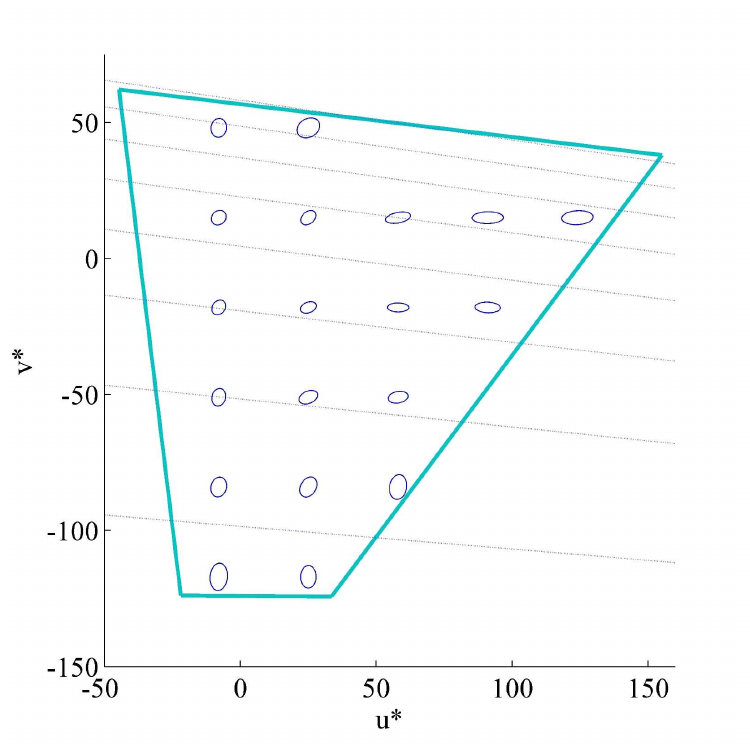}
\end{center}
\caption{Discrimination ellipses of a $D$-Type color-weak observer ($L^*=50$)}
\label{fig:type_d_ellipse_light3}
\end{minipage}
\end{tabular}
\end{figure}

\begin{figure}[htb]
\begin{tabular}{cc}
\begin{minipage}{0.5\textwidth}
\begin{center}
\includegraphics[scale=0.9]{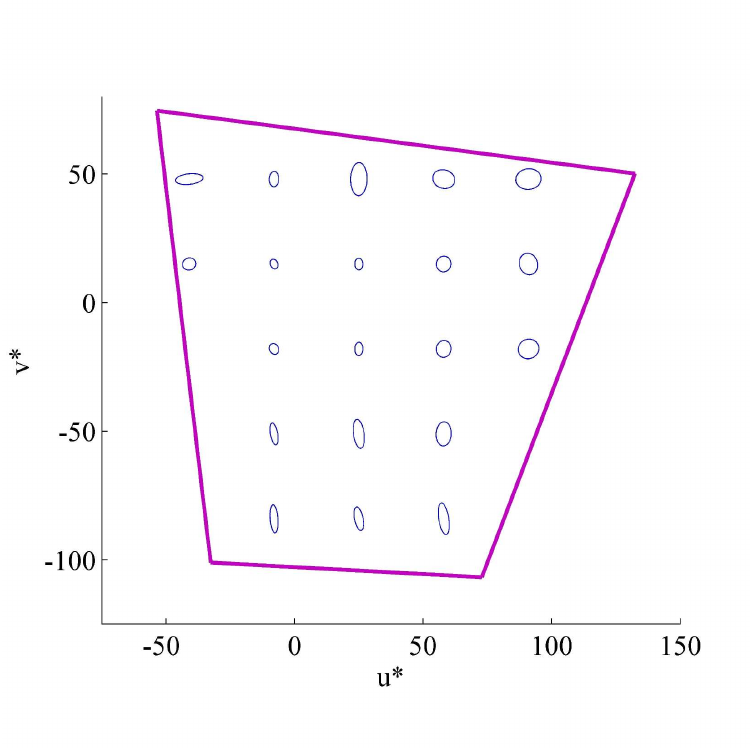}
\end{center}
\caption{Discrimination ellipses of a color-normal observer ($L^*=60$)}
\label{fig:type_c_ellipse_light4}
\end{minipage}\\
\begin{minipage}{0.5\textwidth}
\begin{center}
\includegraphics[width=0.9\columnwidth]{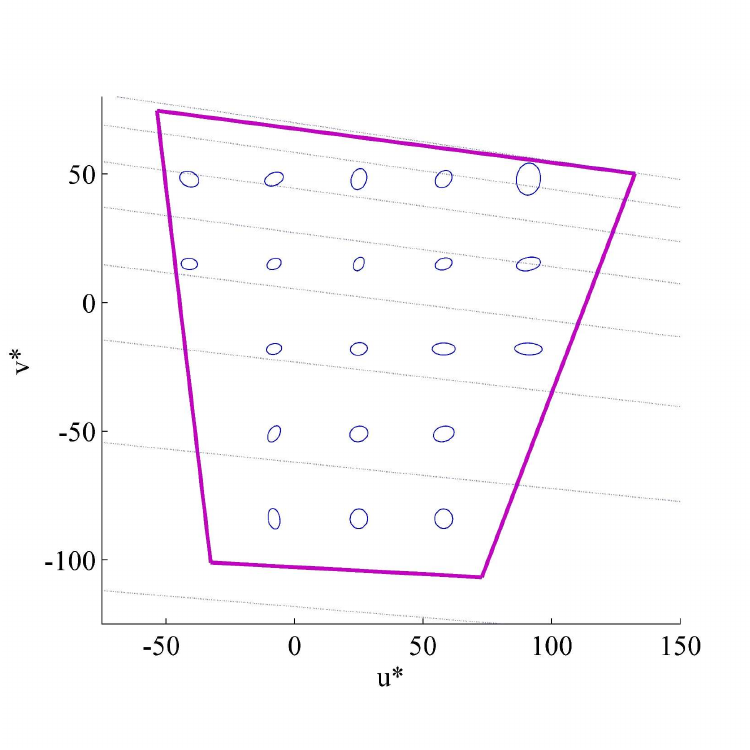}
\end{center}
\caption{Discrimination ellipses of a $D$-Type color-weak observer ($L^*=60$)}
\label{fig:type_d_ellipse_light4}
\end{minipage}
\end{tabular}
\end{figure}

\begin{table}[htb]
\begin{center}
\caption{Area ratio of the ellipsoids between color-normal and color-weak observers}
\label{tab:ellipses_volume}
\begin{tabular}{c|c}
$L^*$ & Area ratio of ellipses\\\hline
$70$ & $2.5172$ \\
$60$ & $1.2074$ \\
$50$ & $1.5590$ \\
$40$ & $0.9675$ \\
$30$ & $2.0069$ \\
Overall & $1.6193$ \\
\end{tabular}
\end{center}
\end{table}


Colors that are not discriminated by an observer with color deficiency of a single cone are located on a line, the so-called confusion line of the type of color-weakness (see~\cite{Color-Science}). In Figures~
\ref{fig:type_c_ellipse_light3}, \ref{fig:type_d_ellipse_light3}, \ref{fig:type_c_ellipse_light4} and~ \ref{fig:type_d_ellipse_light4}
 we show the discrimination ellipses for the L~=~30, 40, 50 and L~=~60~lightness planes of a color-normal observer and the D-type color-weak observer. The frame indicates the gamut at this lightness level while the dashed lines show the confusion lines the D-type color-weak observers. The differences between discrimination thresholds of the color-normal and the color-weak observers can be clearly seen. In Table \ref{tab:ellipses_volume} we collect the area ratio between the discrimination ellipses of the color-normal and the color-weak observers. The values are almost all greater than one which was to be expected. 

At the same time, it is interesting to notice that although the color-weak observer is diagnosed by the Anomaloscope as D-type, the shapes, sizes and orientations of the ellipses vary widely over the color space. In particular, they are not always extended only along the confusion lines of D-type color-weakness or the $M$ direction in the LMS color space which indicates greater insensibility to green for a typical $D$-type observer.
The discrimination threshold ellipsoids actually extend simultaneously also along the confusion lines of $P$-Type and $T$-Type color-weakness or the $L$ and $S$ directions in the LMS color space which suggest greater insensibility to red and blue.
 In fact, this is often found in discrimination threshold measurements  
These deviations of the ellipses from the confusion lines of the ``diagnosed" color-weak type suggest that usually color weakness occurs in more than one type or direction. Color-weakness is thus rarely of a single type but a compound phenomenon. This also means that it is inefficient to describe color-weakness using 1D models like confusion lines. A more precise representation and compensation of color vision characteristics is only possible in higher dimensions. Similarly, one can find that the discrimination threshold ellipses at different lightness levels are also quite different. This can be seen by comparing the results for the levels L~=
50 and L~=~60. These results show that color-weakness is a complex deformation of 3D color space, not only along confussion lines and not only restricted to the chromaticity plane. Practically this also implies that it cannot be exactly described and compensated by a Cartesian product of an 1D model of lightness and a 2D model of the chromaticity plane~\cite{regan}.

\section{Construction of Riemann normal coordinates in color spaces}

Our measurements show that color-weakness is characterized by a complicated deformation of 3D color space. Compensation maps and simulation methods should therefore also use three-dimensional geometry. However, the costs of constructing Riemann normal coordinates in higher dimesional spaces is also high and there is thus a trade-off between accuracy and computational costs. In the following we will describe different approximation methods and study their performance. We will describe methods using only chromaticity, approximations based on the Cartesian product of a one-dimensional model for lightness and a 2D model for chromaticity and full three-dimensional models. For interpolation of the discrimination thresholds, after a Gaussian smoothing, the cubic B-spline and the Akima algorithm \cite{akima} were used to calculate the Christoffel symbols.

An algorithm to build isometries and compensation algorithm in one-dimensional spaces is described in~\cite{CCIW2011}. It was applied to confusion lines for one-dimensional color-weak compensation. Here we use it in the lightness direction to obtain a compensation along the lightness axis. To obtain the Riemann metric on the~$L$ axis, we calculate the intersection of the discrimination ellipsoids centered $l$ on the $L$ axis and the $L$ axis. The resulting interval on the $L$-axis defines the jnd threshold on the $L$ axis. In this way, one obtains the metric tensor $G_n(l)$ for the color-normal and $G_w(l)$ for the color-weak observers.We then apply the algorithm from~\cite{CCIW2011} to match these metrics in the two spaces to obtain a correspondence between the lightness axes of the color-normal and the color-weak observers. This correspondence defines the isometry or color-weak map between the two one-dimensional spaces, which will later be used to build a 2D+1D compensation scheme.

In the case of the two-dimensional chromaticity plane one starts the geodesics from the common origin $(u^*,v^*)=(0,0)$ for each lightness level, which is the same for both color-normal and color-weak observer. The common reference direction given by the zero angle in both, the color spaces of the color-normal and the color-weak observers, is defined as the invariant hue given by the mono-chromatic color with $475\ nm$ \cite{Brettel}.

The next step is to calculate geodesics emanating from the origin, uniformly separated in equal angular increments. This is done using the fourth degree Runge-Kutta algorithm. The unit initial speed is chosen so we actually have the natural parameter as the curve length. The lengths and the angles are measured using metric tensor at the origin. We computed $36$ geodesics from the origin separated by 10~degrees from each other. We also applied a multi-patch strategy here. We thus built an additional coordinate system starting from a second origin which was chosen from a part with low density in the first grid.

\begin{figure}[htb]
\begin{center}
\includegraphics[width=0.9\columnwidth]{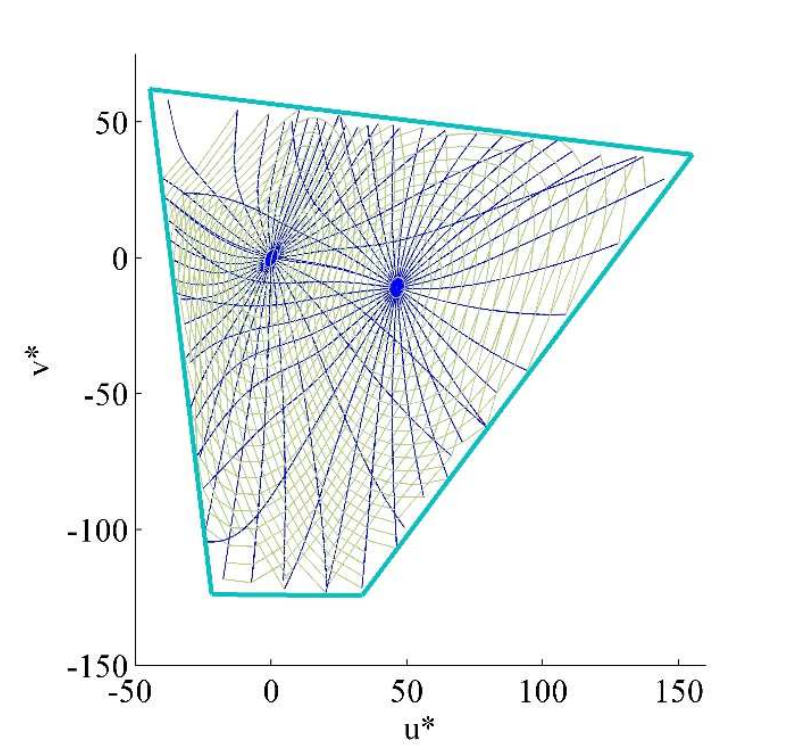}
\end{center}
\caption{Riemann normal coordinates for a color-normal observer ($L^*=50$)}
\label{fig:type_c_2d_rnc_light3}
\begin{center}
\includegraphics[width=0.9\columnwidth]{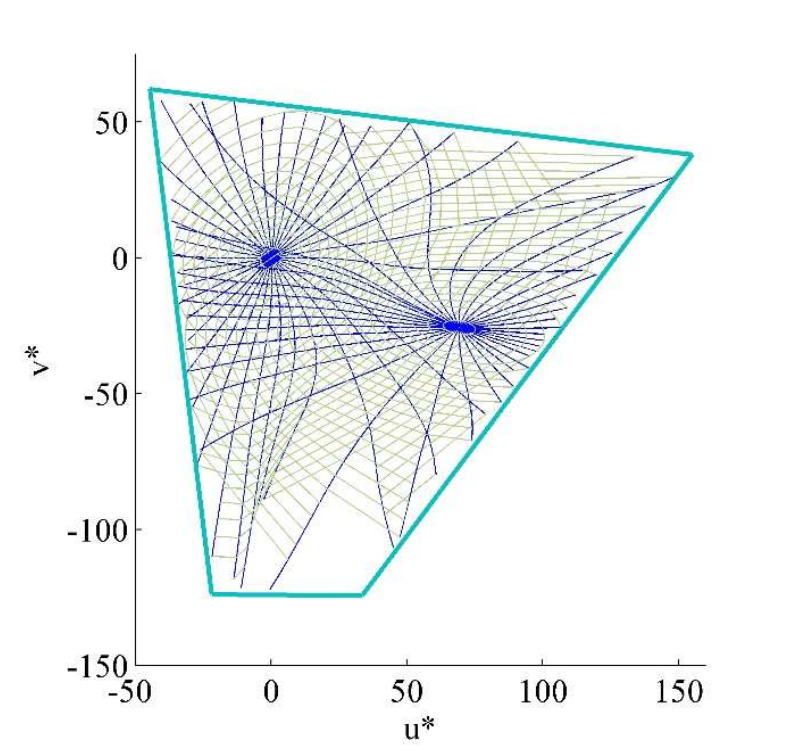}
\end{center}
\caption{Riemann normal coordinates for a $D$-type color-weak observer ($L^*=50$)}
\label{fig:type_d_2d_rnc_light3}
\end{figure}

The Riemann normal coordinates in $L=50, 60$ for normal and color-weak observers are shown in Figs. \ref{fig:type_c_2d_rnc_light3} and \ref{fig:type_d_2d_rnc_light3}. The frames in the figures indicate the gamut at this lightness level. The blue solid lines are geodesics, the yellow-greenish solid lines are equa-chroma lines. Comparing the coordinates at different lightness levels also revealed a distortion along the lightness direction so that the coordinates system is not a Cartesian product between a 1D coordinates and a 2D coordinates. 

We counted the number of the grid points inside the gamut at each lightness level which describe the size of the gamut. The ratios between the number of grid points of the color-weak over the corresponding number of the color-normal observer are shown in the Table~\ref{tab:2d_rnc_volume}. For the color-weak observer this ratio is around $0.7918$ times of that of the color-normal. The grid points in the two color space are separated by an equal distance due to the isometry and we conclude therefore that the color-weak observer has more sparse coordinate grid than the color-normal observer. This agrees with the previous observations about the difference between the two color perception properties.

\begin{figure}[htb]
\begin{center}
\includegraphics[width=0.9\columnwidth]{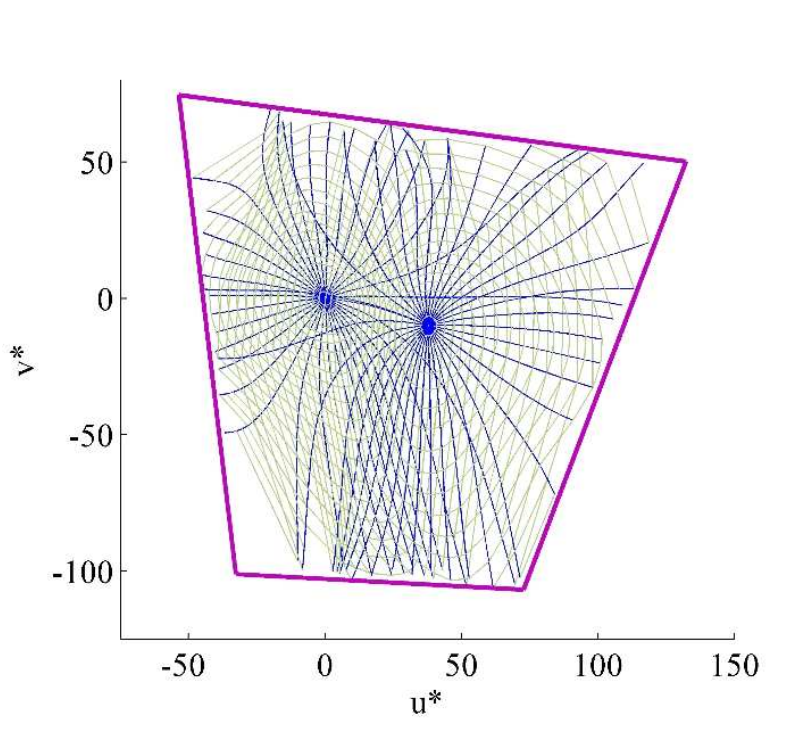}
\end{center}
\caption{Riemann normal coordinates for a color-normal observer ($L^*=60$)}
\label{fig:type_c_2d_rnc_light4}
\end{figure}

\begin{figure}
\begin{center}
\includegraphics[width=0.9\columnwidth]{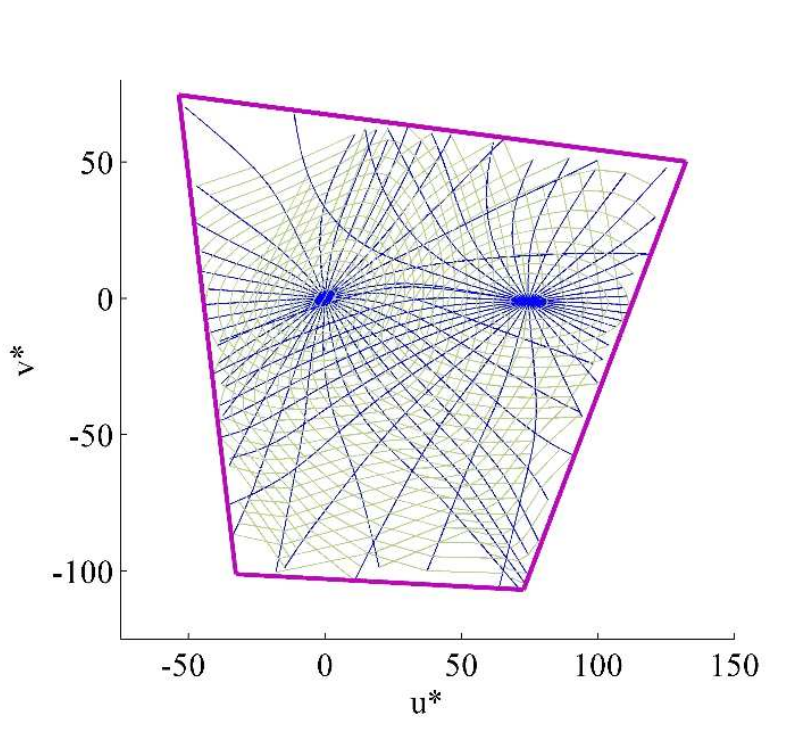}
\end{center}
\caption{Riemann normal coordinates for a $D$-type color-weak observer ($L^*=60$)}
\label{fig:type_d_2d_rnc_light4}
\end{figure}

\begin{table}[htb]
\begin{center}
\caption{Ratio of the number of grid points of color-weak  over color-normal observers (2D)}
\label{tab:2d_rnc_volume}
\begin{tabular}{c|c}
$L^*$ & Ratio of the number of grid points \\\hline
$30$ & $0.7084$ \\
$40$ & $0.9411$ \\
$50$ & $0.7930$ \\
$60$ & $0.8649$ \\
$70$ & $0.6519$ \\
Overall & $0.7918$ \\
\end{tabular}
\end{center}
\end{table}

In the case of the three-dimensional model the common origin for the Riemann normal coordinates can be chosen as the origin on the lowest level of lightness. e.g. in our case, we selected $(L^*,u^*,v^*)=(30,0,0)$. In this case we computed $13 \times 18 = 234$ geodesics emanating from the origin, separated by the same spatial angle and unit initial speed. Distances and angles are again calculated using the metric tensor $G(x)$ at the origin, the common reference direction between the color spaces of color-normal and color-weak observers is again defined by the invariant hue $475\ nm$ in \cite{Brettel}. We also used the multi-patch algorithm in both color spaces.

\begin{figure}[htb]
\begin{center}
\includegraphics[width=0.9\columnwidth]{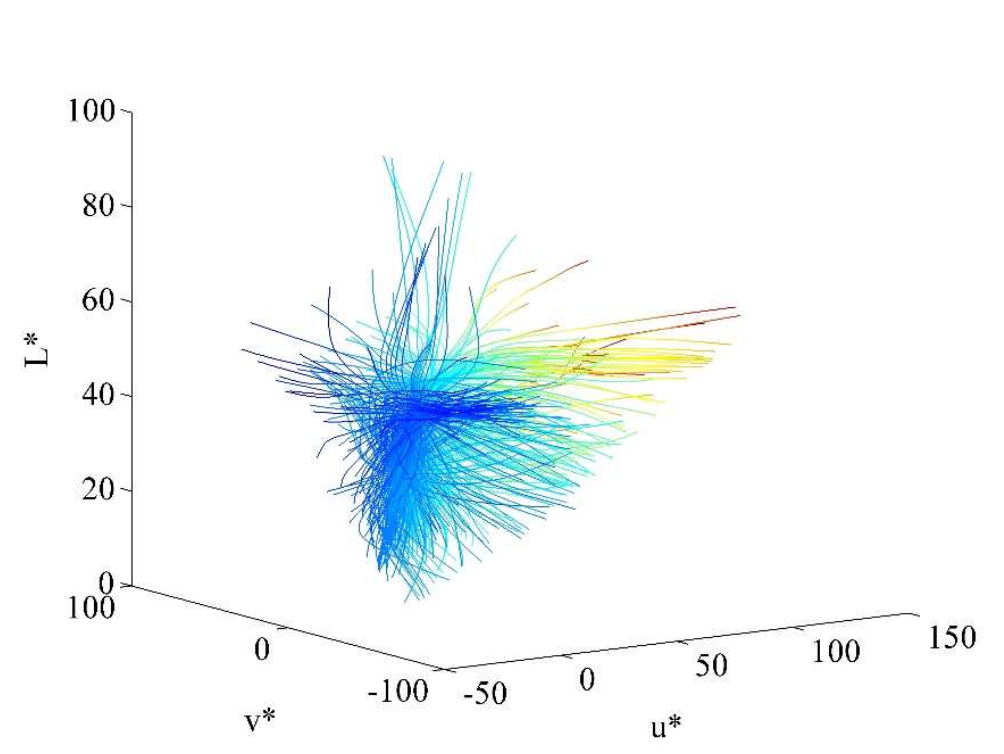}
\end{center}
\caption{Riemann normal coordinates in CIELUV space of the color-normal observer}
\label{fig:type_c_3d_rnc_angle1}

\end{figure}

\begin{figure}[htb]
\begin{center}
\includegraphics[width=0.9\columnwidth]{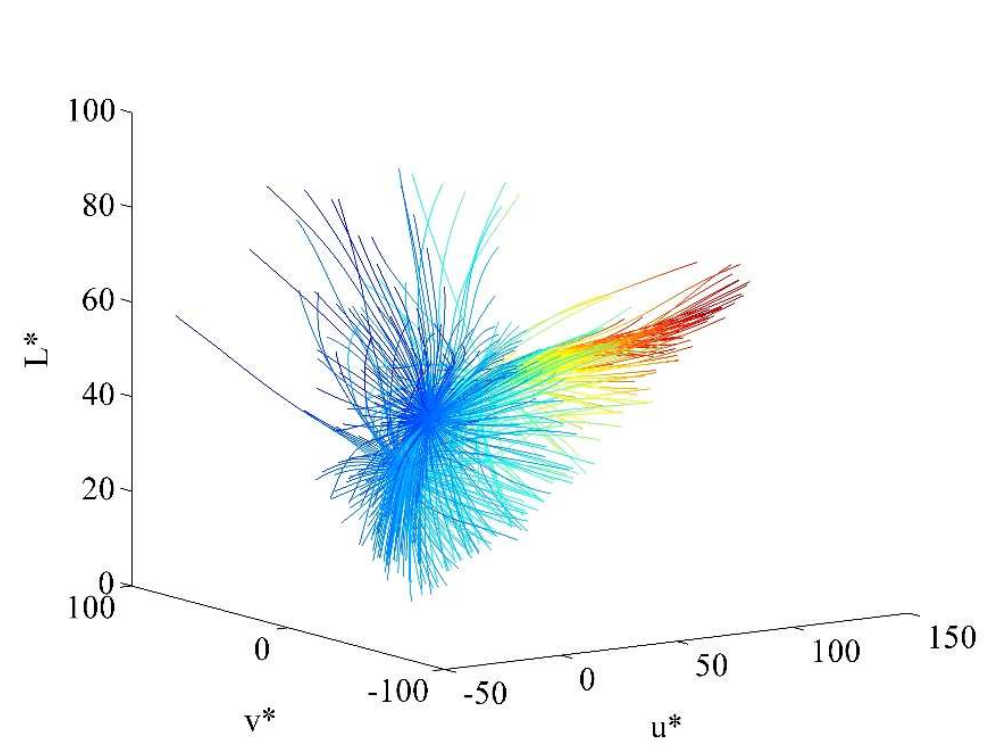}
\end{center}
\caption{Riemann normal coordinates in CIELUV space of the $D$-type color-weak observer}
\label{fig:type_d_3d_rnc_angle1}

\end{figure}

The Riemann normal coordinates in CIELUV space of the color-normal and the color-weak observers are shown in Figs \ref{fig:type_c_3d_rnc_angle1}
and \ref{fig:type_d_3d_rnc_angle1}
.  For better visibility, only geodesics are shown here. The ratio of the number of the grid points inside the gamut for the color-weak over the color-normal observer is $0.6785$, so the difference in the densities of the coordinate grid between the color-normal and the color-weak observers is even greater in the 3D space.

\section{Compensation of natural images}

Using the Riemann normal coordinates in the color spaces of the color-normal and the color-weak observers, one can now construct an isometry between two color spaces using the approach described previously. We will now describe the three implementations of the color-weak compensation for the $2D$, the $2D+1D$ and the $3D$ case.

\subsection{2D case}

\begin{description}
\item[Step 1]
For the chromaticity planes with fixed lightness level $l$ in the color-normal and color-weak observers, 
build the compensation map  $\gamma_l:C_n\longrightarrow C_w$ as the inverse map of the color-weak map $w_l$ for each $l$; 

\item[Step 2]
Transform the $RGB$ coordinates of an input image to CIELUV coordinates in $C_n$;

\item[Step 3]
Choose the $l$ which is the closest to $L^*$, and choose the compensation map as $\gamma_l$;

\item[Step 4]
From the grid of the Riemann normal coordinates of $C_n$, find the triangular patch which contains $(u^*,v^*)$. The Riemann normal coordinates of the input color is taken as e.g. either that of the closest vertex $(u_0^*,v_0^*)$, or a convex combination of the vertices of the triangel $(u_i^*, v_i^*), i=1, ..., 3$, i.e., 
\[(u^*, v^*)=\sum_i a_i (u_i^*, v_i^*), a_i\geq 0, \sum_i a_i=1\] 

\item[Step 5] Apply the map $\gamma_l$ to the input color, e.g., the vertex $(u_0^*,v_0^*)$ in $C_n$ becomes the vertex $(u_0^{*'},v_0^{*'})$ in $C_w$, or, assuming that $\gamma_l$ is a linear map of the simplex, then the vertices  $(u_i^*, v_i^*), i=1, ..., 3$ become  $(u_i^{*',} v_i^{*'}), i=1, ..., 3$, and we get:  
\begin{eqnarray}
(u^{*'},v^{*'})&=&\gamma_l((u^*, v^*))\\
&=&\sum_i a_i \gamma_l((u_i^*, v_i^*))\\
&=&\sum_i a_i(u_i^{*',} v_i^{*'})
\end{eqnarray}

\item[Step 6]
Repeat Step 3 to Step 6 for all pixels in the input image;

\item[Step 7]
Transform the CIELUV coordinates to $R'G'B'$ coordinates
\end{description}

\subsection{2D +1D case}

The 2D algorithm does not take into account the distortion of color-weak vision along the lightness direction. Next we add to the 2D algorithm a 1D compensation along the lightness direction. The 1D compensation algorithm described in the previous section is then applied to the $L$ axis. The algorithm is then exactly the same as in the previous section but with a new step to compensate the $l$ values simultaneously.

\subsection{3D case}%
For the full 3D case we get finally the following processing steps:
\begin{description}
\item[Step 1]

For color spaces $C_n , C_w$ of the color-normal and color-weak observers, 
build the compensation map  $\gamma:C_n\longrightarrow C_w$ as the inverse map of the color-weak map $w$; 

\item[Step 2]
Transform the $RGB$ coordinates of an input image to CIELUV coordinates in $C_n$;

\item[Step 3]
From the grid of the Riemann normal coordinates of $C_n$, find the tetrahedron which contains $(L^*,u^*,v^*)$. The Riemann normal coordinates of the input color is taken as e.g. either that of the closest vertex $(L_0^*, u_0^*,v_0^*)$, or  a convex combination of the vertices of the tetrahedron $(L_i^*, u_i^*, v_i^*), i=1, ..., 4$, i.e., 
\[(L^*, u^*, v^*)=\sum_i a_i (L_i^*, u_i^*, v_i^*), a_i\geq 0, \sum_i a_i=1\] 

\item[Step 4] Apply the map $\gamma$ to the input color, e.g., the vertex $(L_0^*, u_0^*,v_0^*)$ in $C_n$ becomes the vertex $(L_0^{*'}, u_0^{*'},v_0^{*'})$ in $C_w$, or , assuming that $\gamma$ is linear on the tetrahedron, the vertices  $(L_i^{*}, u_i^*, v_i^*), i=1, ..., 4$ become  $(L_i^{*'}, u_i^{*',} v_i^{*'}), i=1, ..., 4$, then  
\begin{eqnarray}
(L^{*'}, u^{*'},v^{*'})&=&\gamma((L^*, u^*, v^*))\\
&=&\sum_i a_i \gamma((L_i^*, u_i^*, v_i^*))\\
&=&\sum_i a_i(L_i^{*'}, u_i^{*',} v_i^{*'})
\end{eqnarray}

\item[Step 5]
Repeat from Step 3 to Step 5 for all pixels in the input image;

\item[Step 6]
Transform the CIELUV coordinates to $R'G'B'$ coordinates
\end{description}

\section{Evaluation}

Since a direct evaluation of the performance of compensation is difficult, especially for natural images with complicated color distributions, we used the semantic differential method~\cite{SD_method} to compare visual impression before and after compensation. The SD method is known as an effective tool for qualitative evaluation of subjective impression by asking an subject to judge between two antonyms for different adjectives. 

We selected eight pairs of adjectives from the Osgood's original 76 pairs. These pairs are relevant for judging visual impressions and they avoid adaptation related choices such as "natural or not", "friendly or not".

In one SD-evaluation the current compensation image is calculated using one of the three compensation algorithms and shown to the color-normal and color-weak observers in a random order. Two series of images showing images called "Lake" and "Pond", are shown as examples. After each presentation the observer completes the SD questionnaire. The results and the correlation coefficients between SD scores of the color normal observer seeing the original and the  color-weak observer seeing the compensated images are shown in Table~\ref{tab:sd_score_coc}. 

In our evaluation the compensation algorithms are applied to ten natural images. The SD scores obtained are shown in the following four cases, "original" and "2D", "2D+1D", "3D" stand for resulting image by 2D, 2D+1D and 3D  compensation algorithms, and color-weak "simulation" of the input image is also shown. For the 30~compensation images, 21~resulted in an increment of the correlation coefficients between the color-normal and the color-weak evaluation. This indicates that the compensation leads to a closer similarity between the perception of the color-weak and the normal observer.

\begin{table}[htb]
\begin{center}
\caption{ Correlation coefficient of SD score}
\label{tab:sd_score_coc}
\begin{tabular}{l|cccc}
Image & original & 2D & 2D+1D & 3D \\\hline
"Lake" & $0.4845$ & $0.6204$ & $0.8518$ & $0.7883$ \\
"Pond" & $-0.5078$ & $-0.0265$ & $-0.2136$ & $0.1039$ \\
\end{tabular}
\end{center}
\end{table}

\begin{figure}[htb]
\begin{center}
\includegraphics[width=0.9\columnwidth]{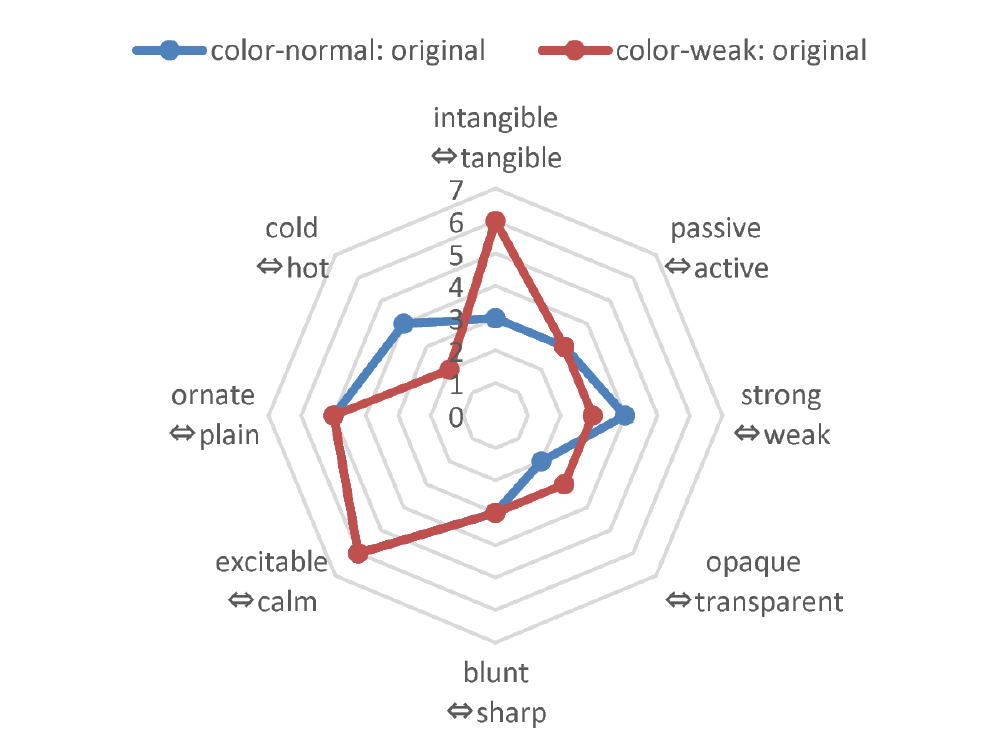}
\end{center}
\caption{"Lake":\newline Color-normal: original  D-type color-weak: original}
\label{fig:sd_lake_c_org_d_org}
\end{figure}
\begin{figure}[htb]
\begin{center}
\includegraphics[width=0.9\columnwidth]{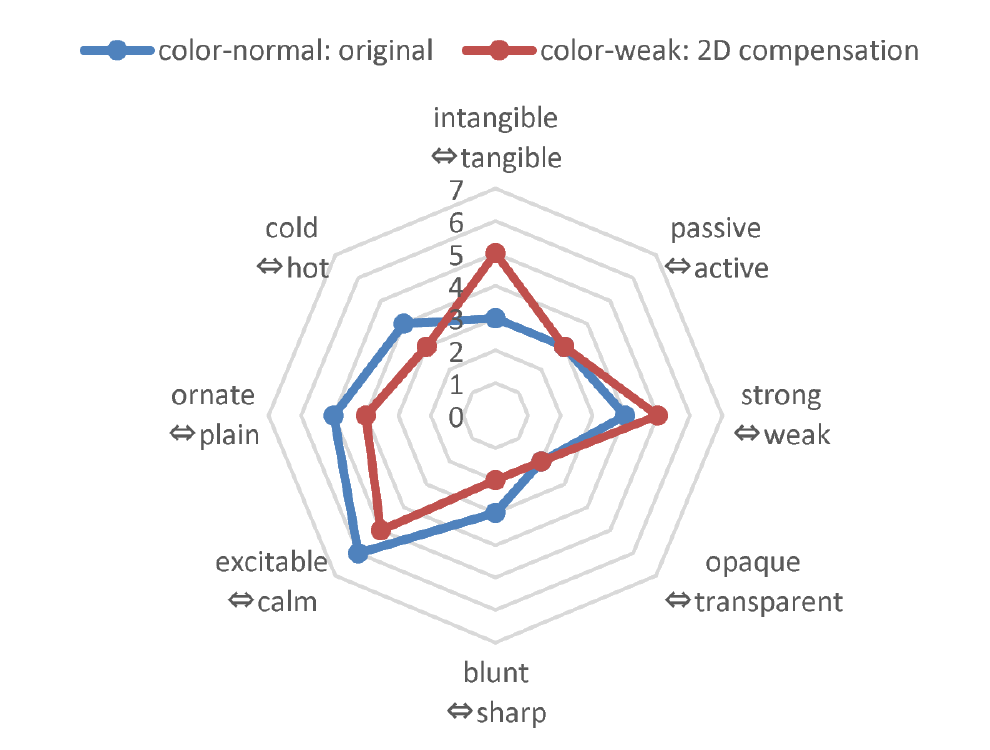}
\end{center}
\caption{"Lake"\newline Color-normal: original  D-type color-weak: 2D compensation}
\label{fig:sd_lake_c_org_d_2dc}
\end{figure}

\begin{figure}[htb]
\begin{center}
\includegraphics[width=0.9\columnwidth]{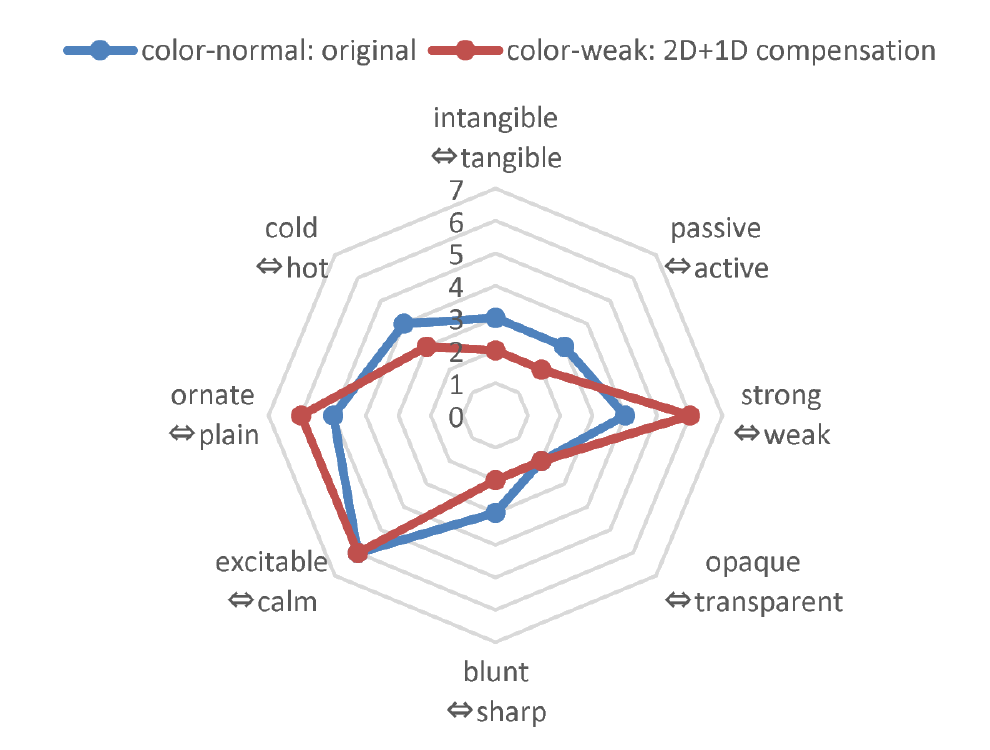}
\end{center}
\caption{"Lake":\newline Color-normal: original D-type color-weak: 2D+1D compensation}
\label{fig:sd_lake_c_org_d_2d1dc}
\end{figure}
\begin{figure}[htb]
\begin{center}
\includegraphics[width=0.9\columnwidth]{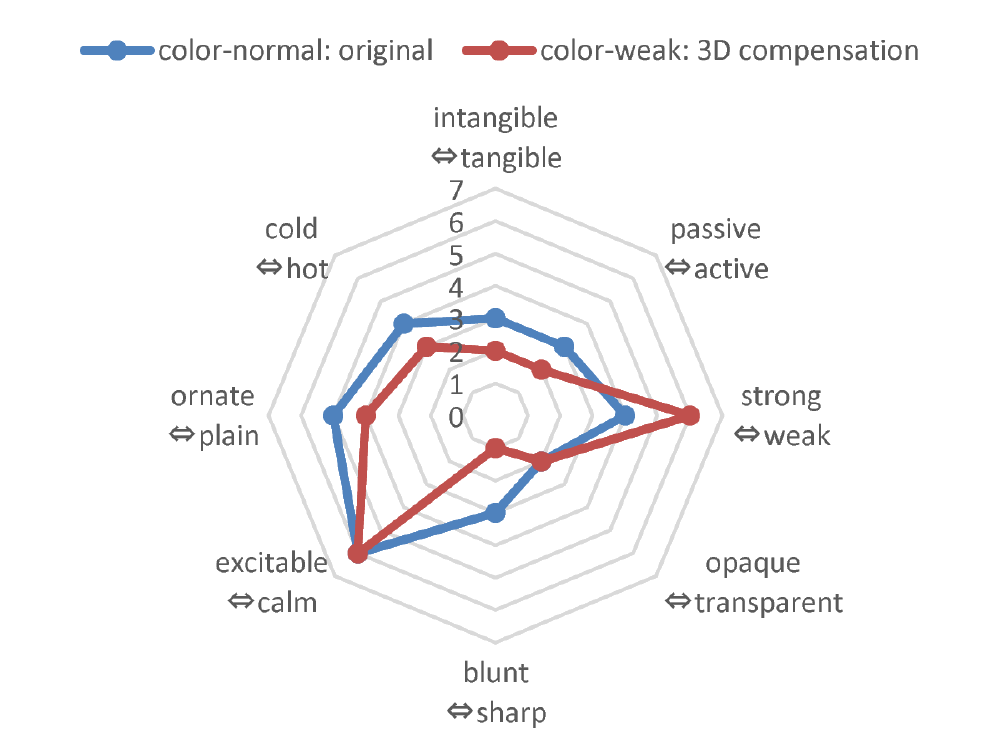}
\end{center}
\caption{"Lake":\newline Color-normal: original D-type color-weak: 3D compensation}
\label{fig:sd_lake_c_org_d_3dc}
\end{figure}

\begin{figure}[htb]
\begin{center}
\includegraphics[width=0.9\columnwidth]{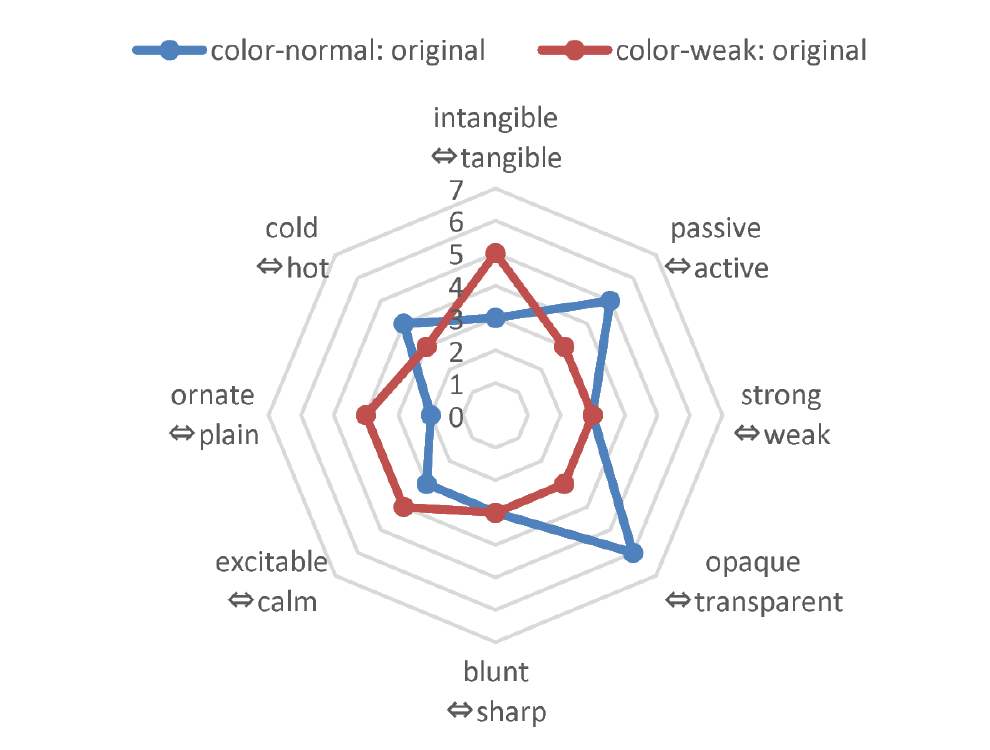}
\end{center}
\caption{"Pond": \newline Color-normal: original D-type color-weak: original}
\label{fig:sd_ike_c_org_d_org}
\end{figure}
\begin{figure}[htb]
\begin{center}
\includegraphics[width=0.9\columnwidth]{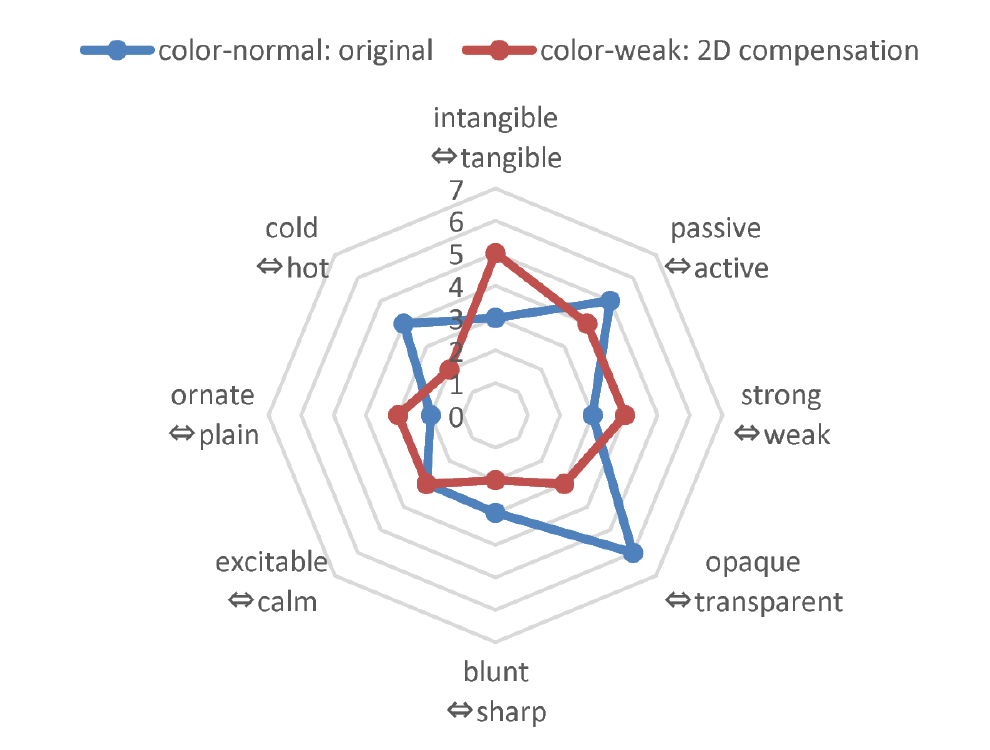}
\end{center}
\caption{"Pond":\newline Color-normal: original  D-type color-weak: 2D compensation}
\label{fig:sd_ike_c_org_d_2dc}
\end{figure}

\begin{figure}[htb]
\begin{center}
\includegraphics[width=0.9\columnwidth]{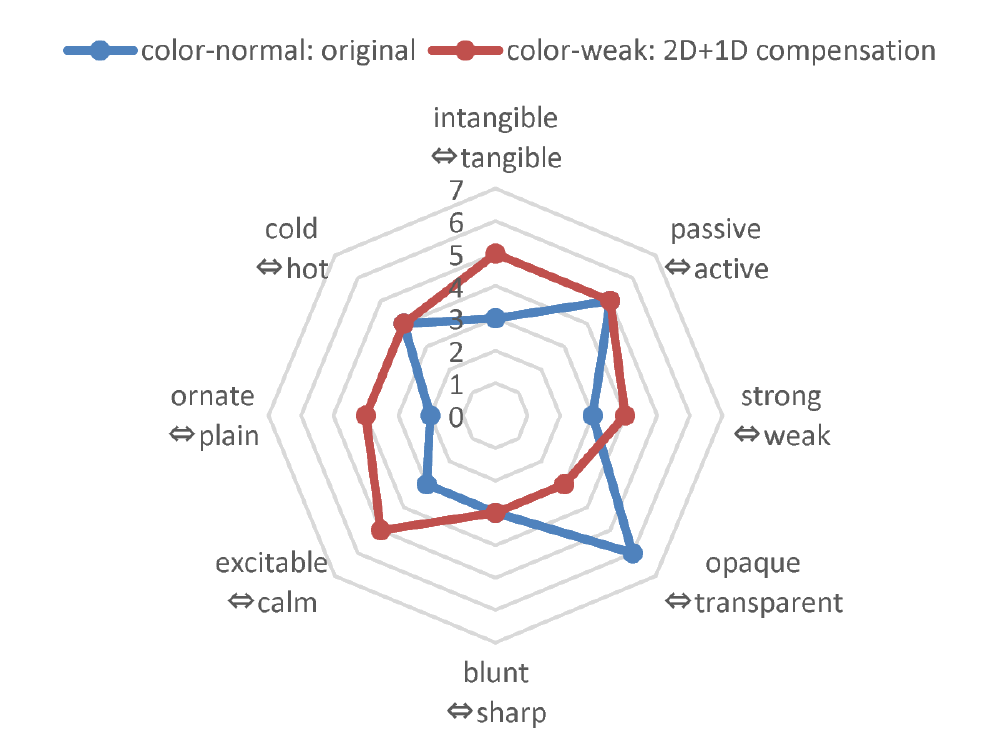}
\end{center}
\caption{"Pond"\newline Color-normal: original  D-type color-weak: 2D+1D compensation }
\label{fig:sd_ike_c_org_d_2d1dc}
\end{figure}
\begin{figure}[htb]
\begin{center}
\includegraphics[width=0.9\columnwidth]{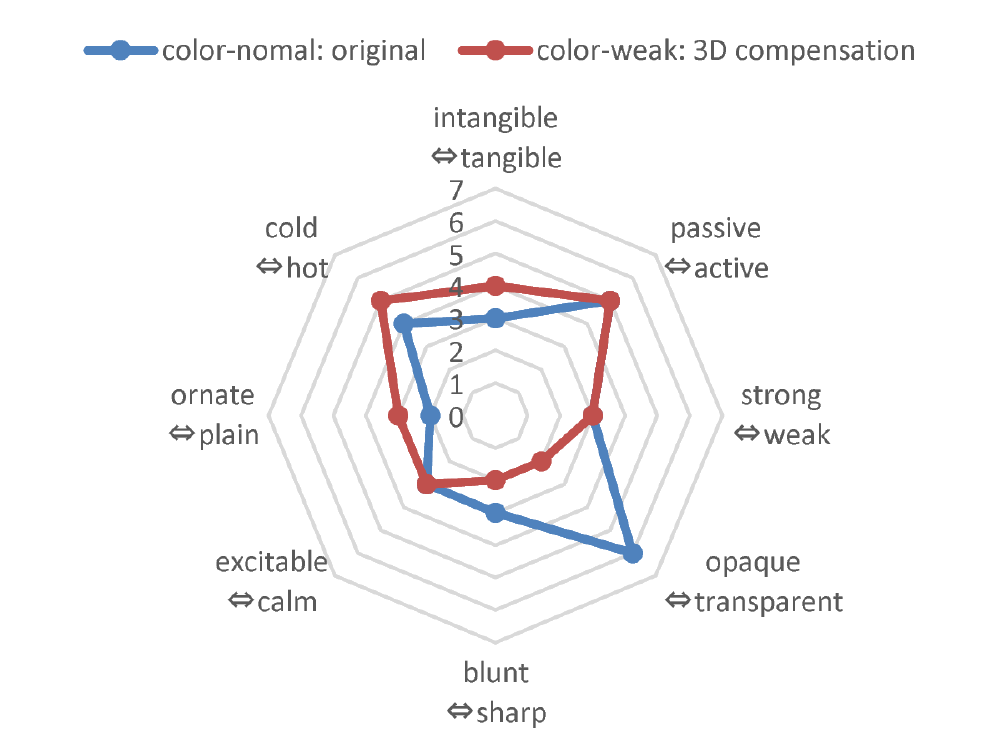}
\end{center}
\caption{"Pond"\newline Color-normal: original  D-type color-weak: 3D compensation}
\label{fig:sd_ike_c_org_d_3dc}
\end{figure}

\section{Compensation results}

The compensation algorithms are applied to ten natural images. The results are shown in the following four cases, "original" and "2D", "2D+1D", "3D" stand for resulting images by 2D, 2D+1D and 3D  compensation algorithms, and color-weak "simulation" of the input image is also shown.

In particular, we show two images called "Lake" and "Pond" as the examples.

Table \ref{tab:image_pixels_volume} shows the area expansion of the pixel distribution on the $u^*v^*$ chromaticity plane after compensation.
The average CIELUV coordinate values are shown in  Table \ref{tab:image_pixels_diff}.

\begin{table}[htb]
\begin{center}
\caption{Area expansion of $u^*v^:$ distribution}
\label{tab:image_pixels_volume}
\begin{tabular}{l|ccc}
Compensation & "Lake" & "Pond" \\\hline
2D        & $1.8182$ & $1.4839$ \\
2D+1D & $1.4596$ & $1.5483$ \\
3D        & $1.4017$ & $1.4765$ 
\end{tabular}
\end{center}
\end{table}

\begin{table}[htb]
\begin{center}
\caption{Averages of CIELUV values}
\label{tab:image_pixels_diff}
\begin{tabular}{ll|ccc}
Image & Algorithm & $L^*$ & $u^*$ & $v^*$ \\\hline
"Lake" & original       & $22.8666$ & $13.0871$ & $38.3313$ \\
 & 2D        & $22.5367$ & $22.1902$ & $38.6488$ \\
 & 2D+1D & $27.5266$ & $12.9285$ & $29.9181$ \\
 & 3D          & $25.0195$ & $16.0550$ & $33.2527$ \\
"Pond" & original          & $19.3940$ & $24.6891$ & $32.6535$ \\
 & 2D        & $18.2399$ & $36.9779$ & $31.9029$ \\
 & 2D+1D & $22.8467$ & $21.6778$ & $29.3835$ \\
 & 3D          & $20.3572$ & $30.2214$ & $27.6951$
\end{tabular}
\end{center}
\end{table}

Generally the distribution of the pixels in the color space is expanded after compensation.
In particular,  the chromaticity distribution of the input image has been expanded after compensation.
Besides, almost every compensation image has large $u^*v^*$ average value. After the lightness compensation, the $L^*$ average value increased.
On the other hand, in the 2D+1D compensation, average of $u^*$ and  $v^*$ decreased, which is due to the gamut shrinking in the higher lightness level.

The average $L^*$ value is larger  in 2D+1D compensation than in 3D compensation, which is because the former expands the lightness the same rate for every $u^*v^*$ which could be too strong,  but the latter takes into account of properties for each chromaticity. It is therefore more flexible.
In fact, 
it can be observed that the 2D compensation enhanced both red and green but with quite different L values.
Meanwhile, the 3D compensation combined chromaticity with lightness to produce more natural tone and contrast and therefore a balanced color distribution.

\begin{figure}[htb]
\centering
\includegraphics[width=0.9\columnwidth]{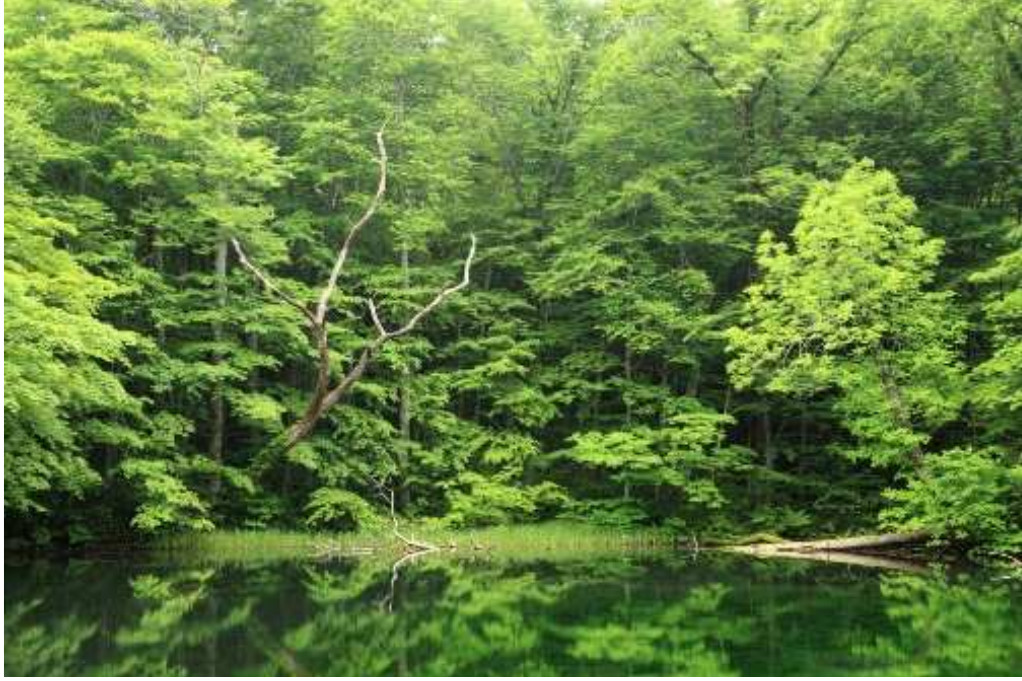}
\caption{"Lake": Original}
\label{fig:pic_lake_org}
\end{figure}

\begin{figure*}
\centering
\subfloat["Lake": 2D compensation]{\includegraphics[width=0.9\columnwidth]{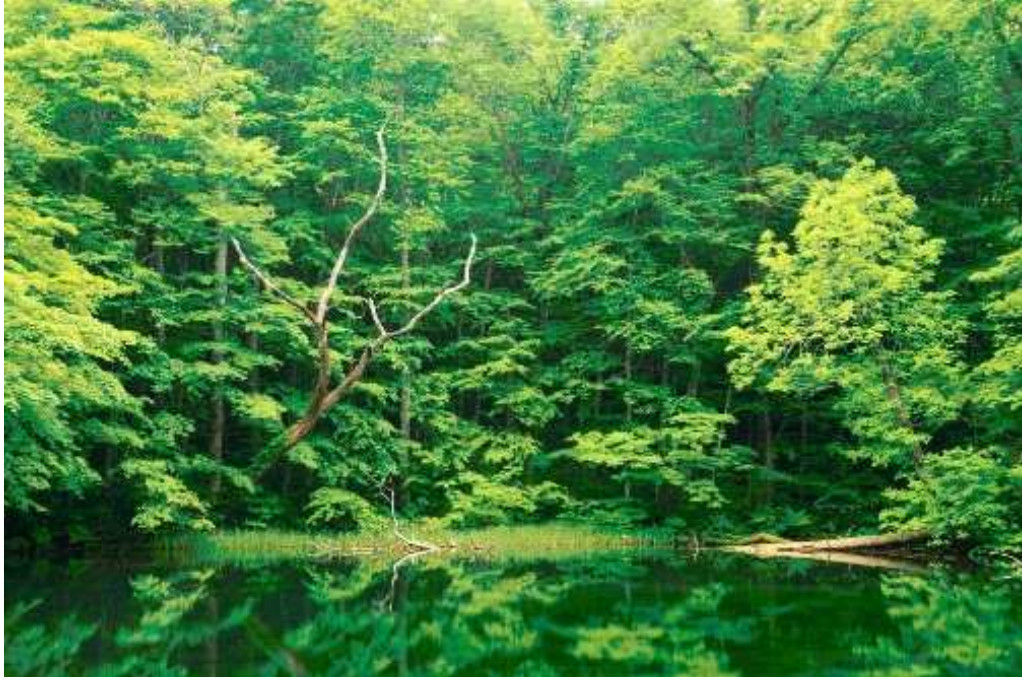}}
\hspace{1cm}
\subfloat["Lake": 2D simulation]{\includegraphics[width=0.9\columnwidth]{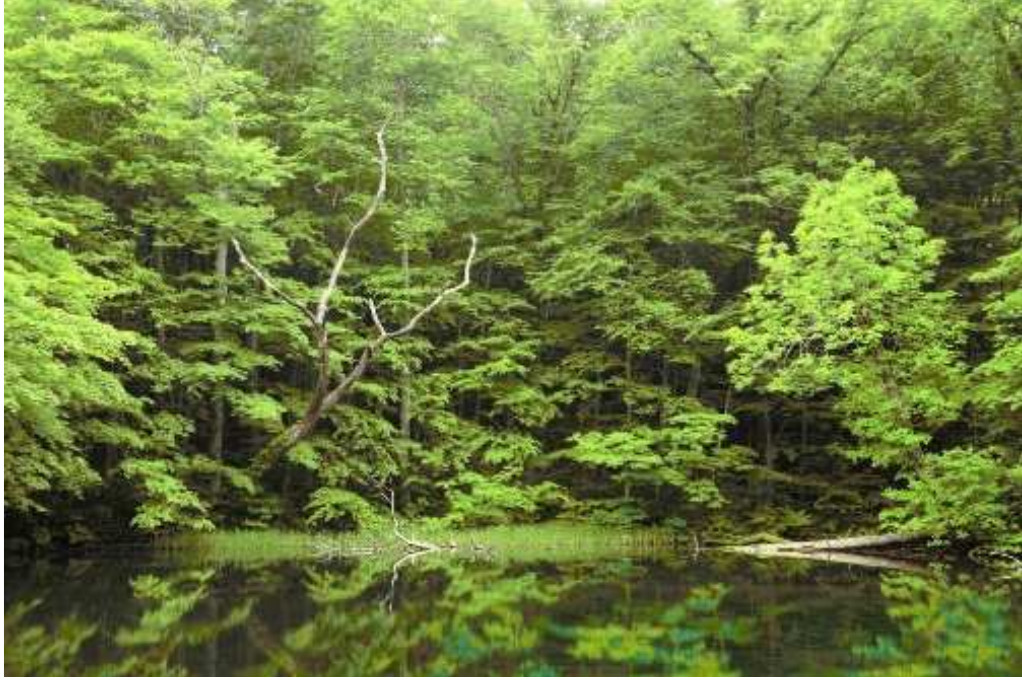}}\\
\subfloat["Lake": 2D+1 compensation]{\includegraphics[width=0.9\columnwidth]{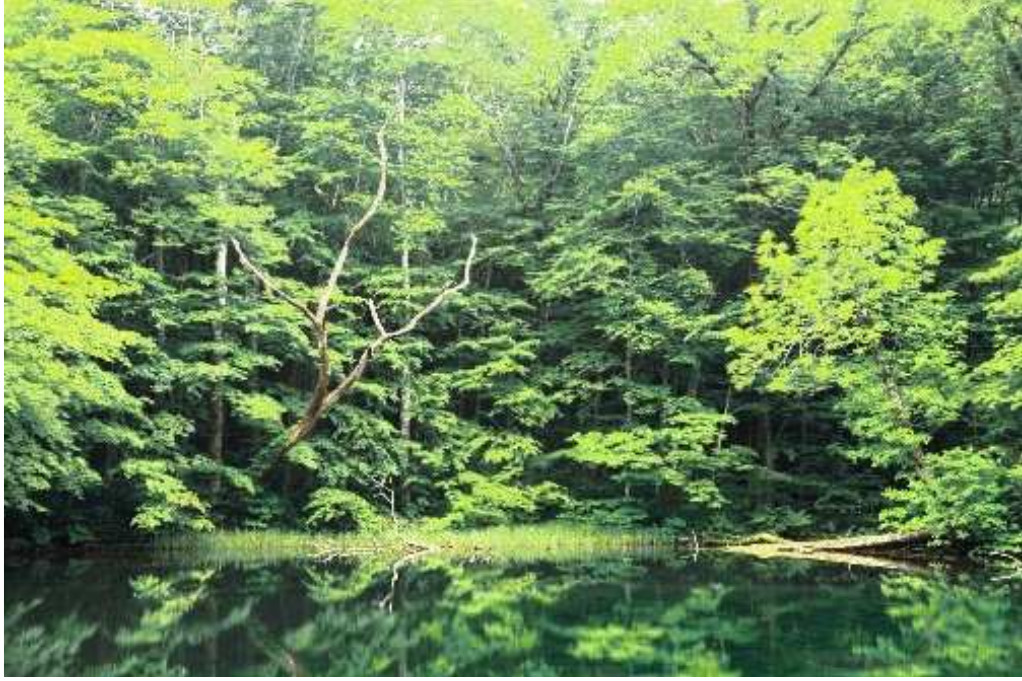}
}
\hspace{1cm}
\subfloat["Lake": 2D+1 simulation]{\includegraphics[width=0.9\columnwidth]{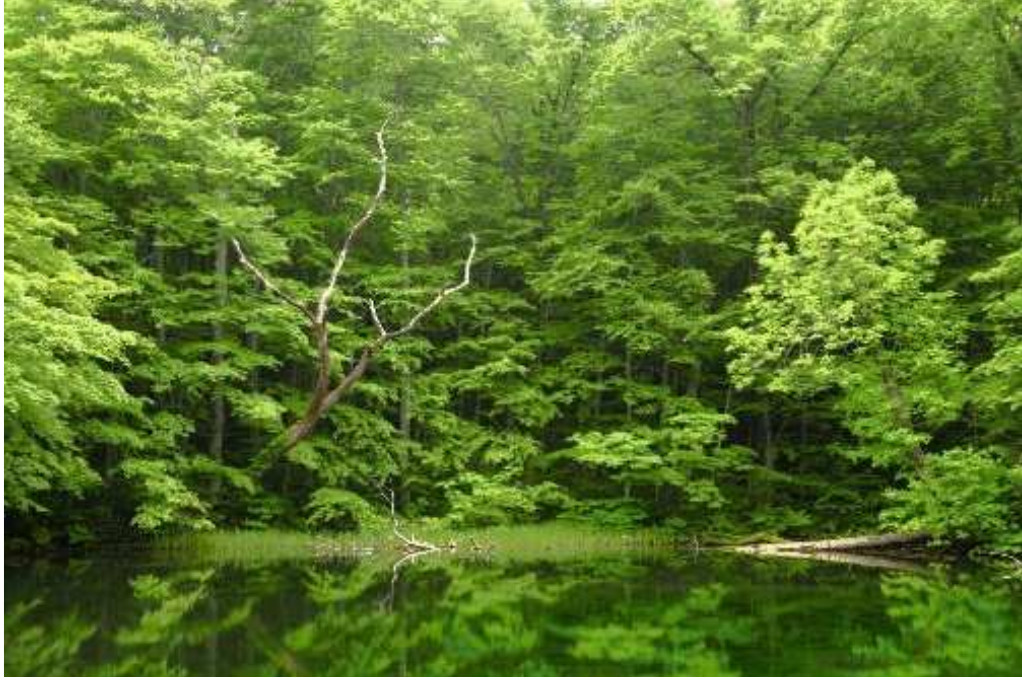}
}\\
\subfloat["Lake": 3D compensation]{\includegraphics[width=0.9\columnwidth]{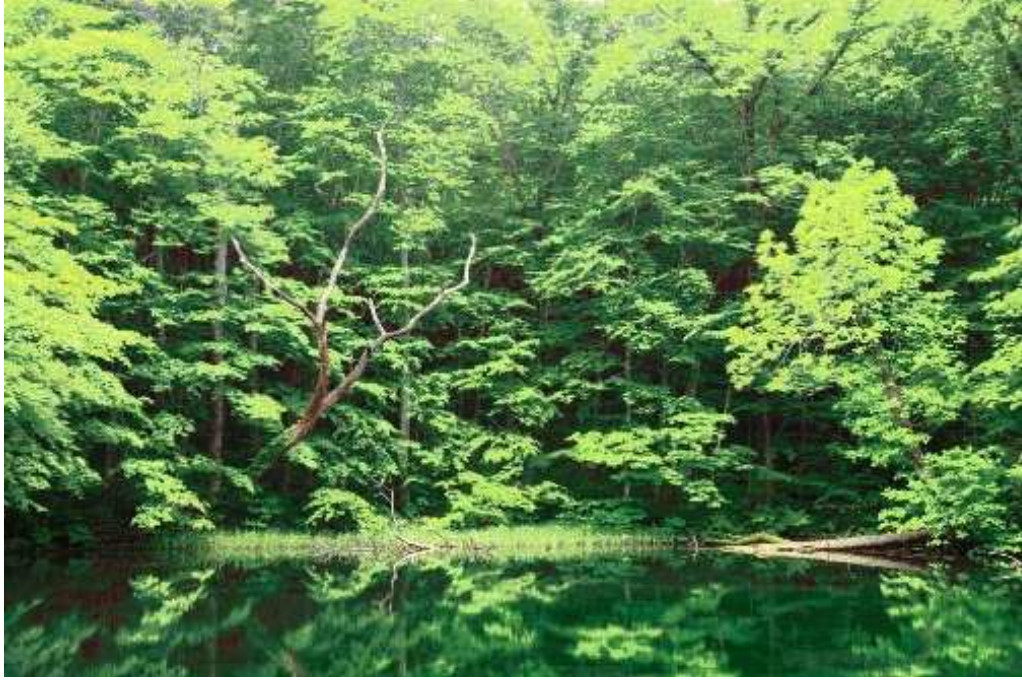}
}
\hspace{1cm}
\subfloat["Lake": 3D simulation]{\includegraphics[width=0.9\columnwidth]{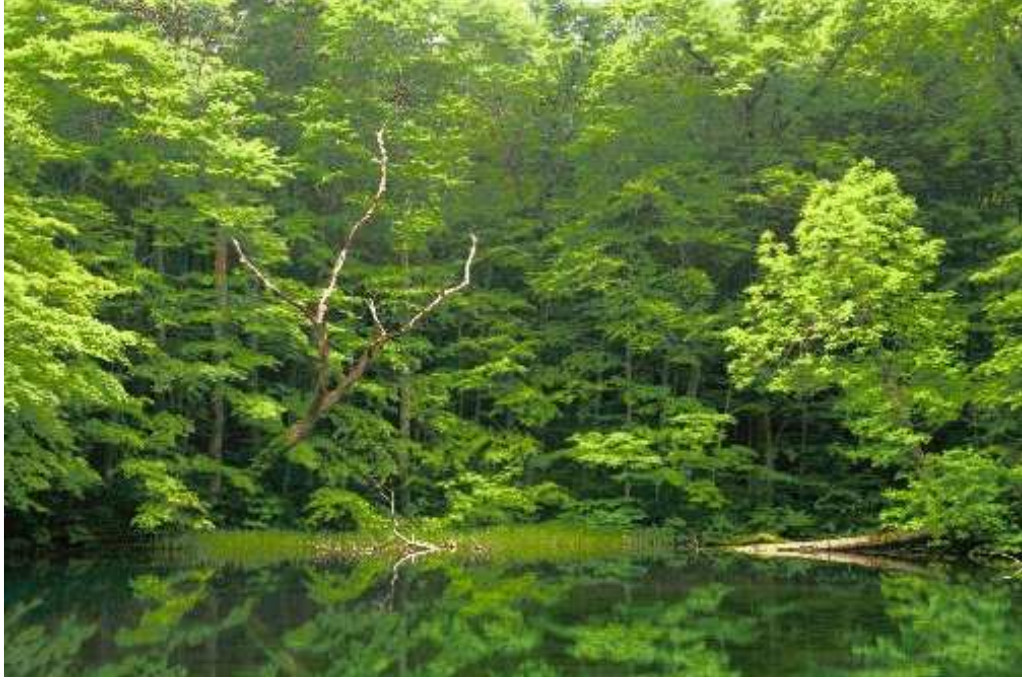}
}
\end{figure*}

\begin{figure}[htb]
\centering
\includegraphics[width=0.9\columnwidth]{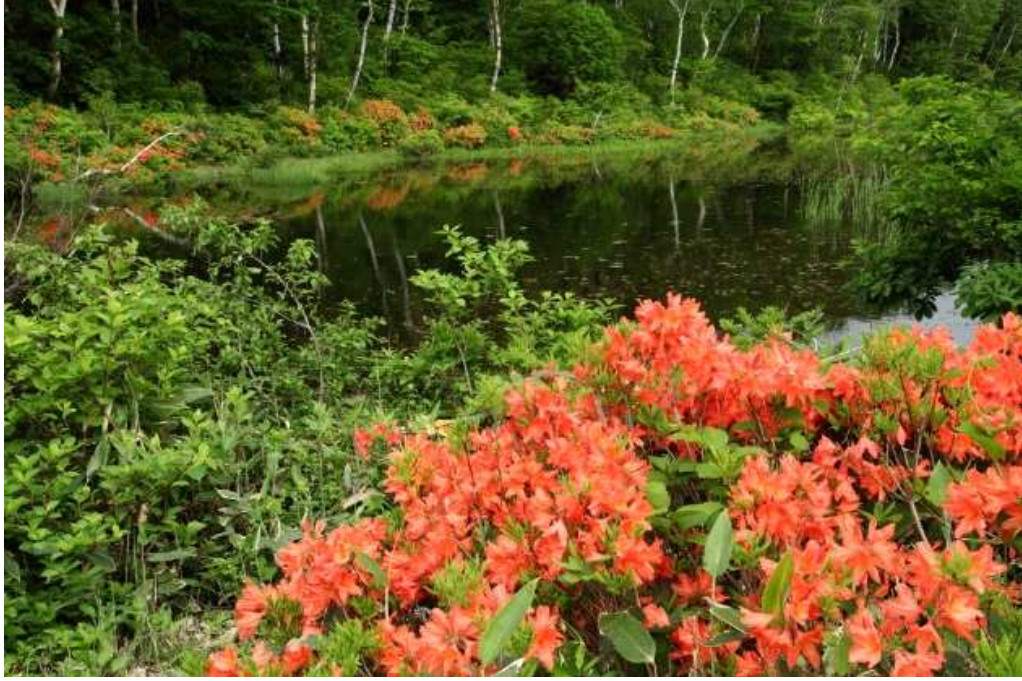}
\caption{"Pond": Original}
\label{fig:pic_ike_org}
\end{figure}

\begin{figure*}
\centering
\subfloat["Pond": 2D compensation]{\includegraphics[width=0.9\columnwidth]{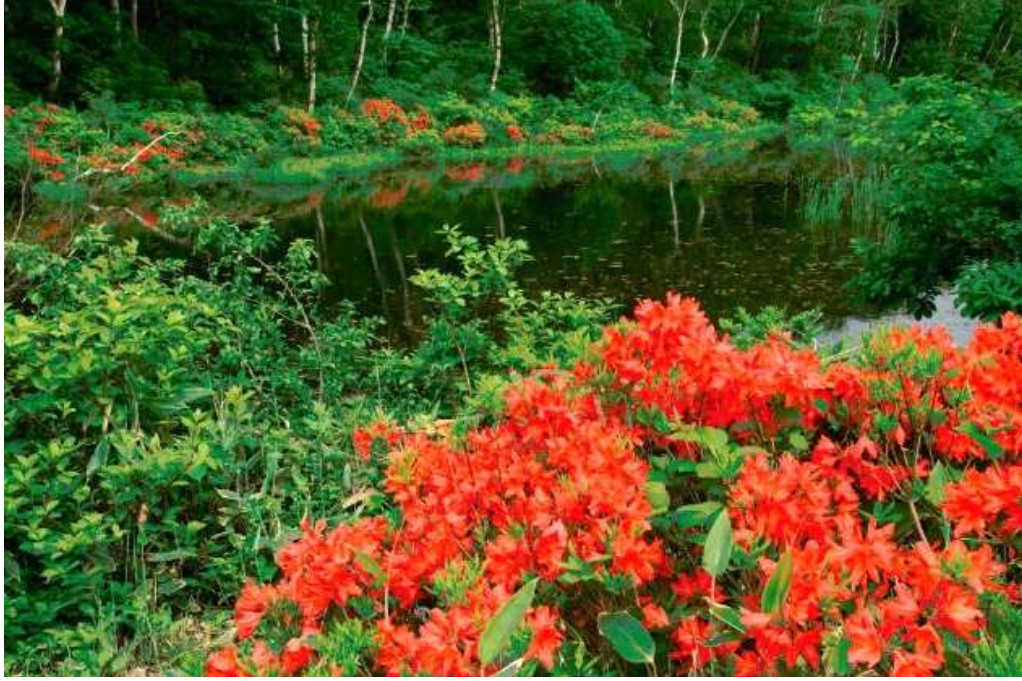}}
\hspace{1cm}
\subfloat["Pond": 2D simulation]{\includegraphics[width=0.9\columnwidth]{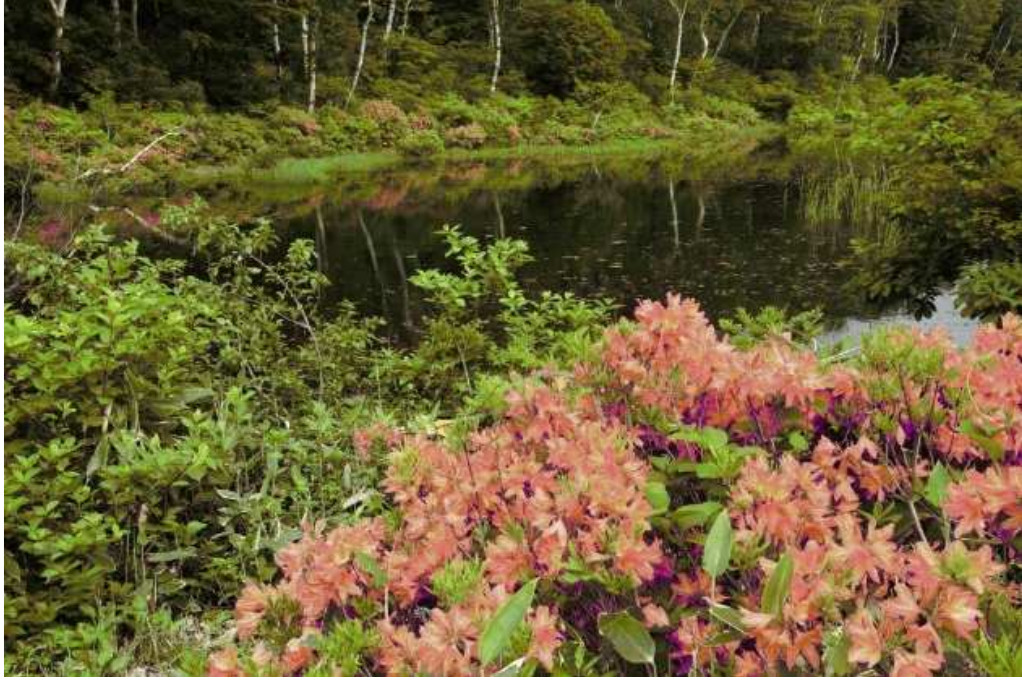}}\\
\subfloat["Pond": 2D+1 compensation]{\includegraphics[width=0.9\columnwidth]{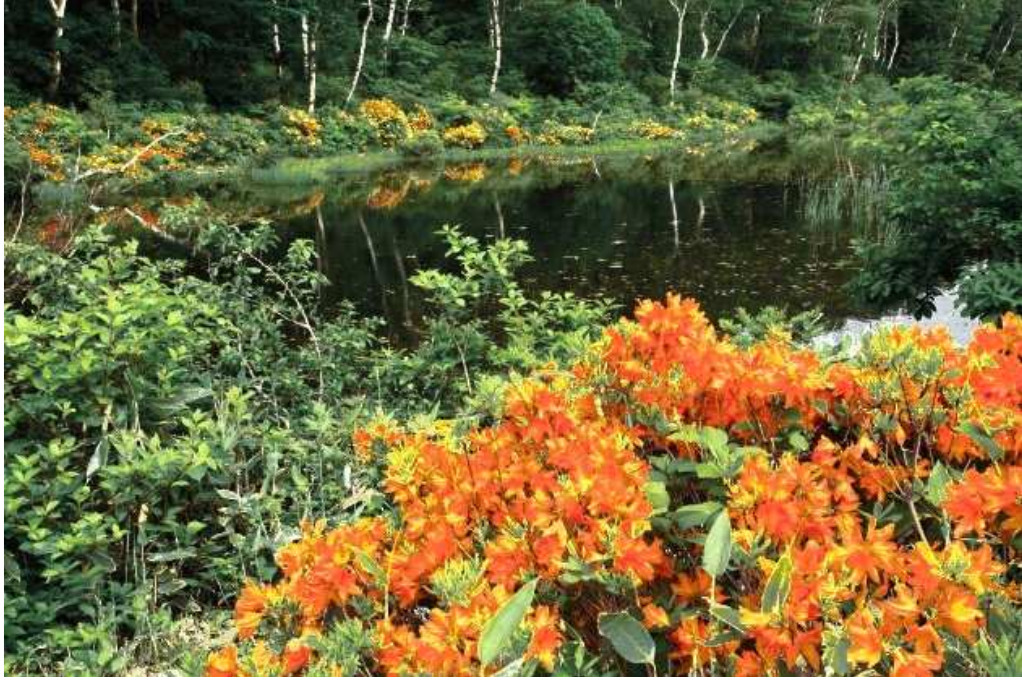}
}
\hspace{1cm}
\subfloat["Pond": 2D+1 simulation]{\includegraphics[width=0.9\columnwidth]{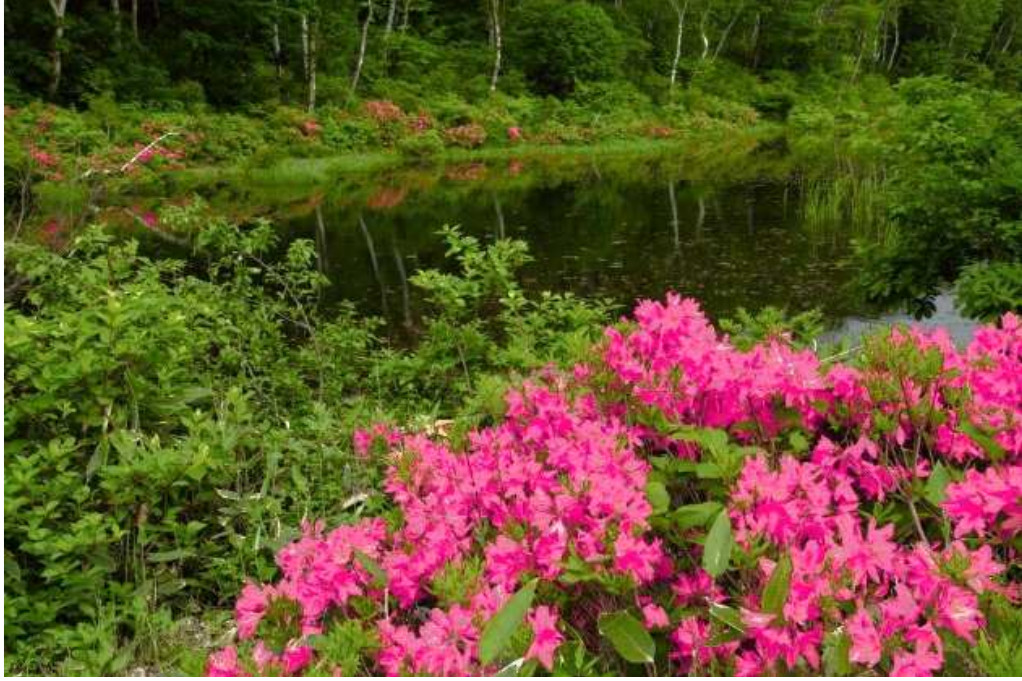}
}\\
\subfloat["Pond": 3D compensation]{\includegraphics[width=0.9\columnwidth]{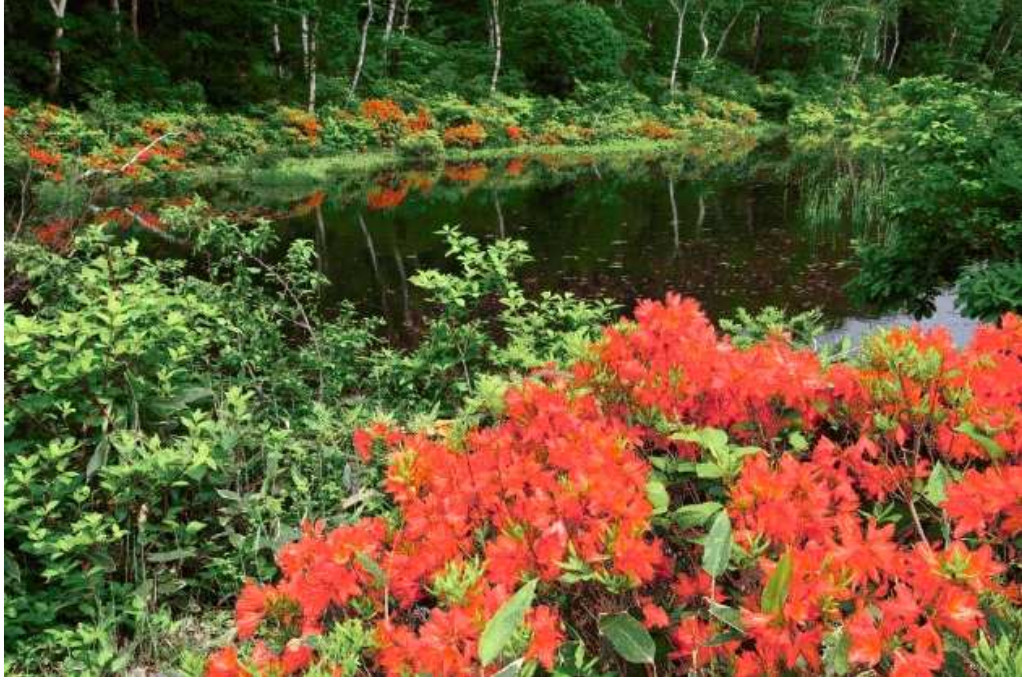}
}
\hspace{1cm}
\subfloat["Pond": 3D simulation]{\includegraphics[width=0.9\columnwidth]{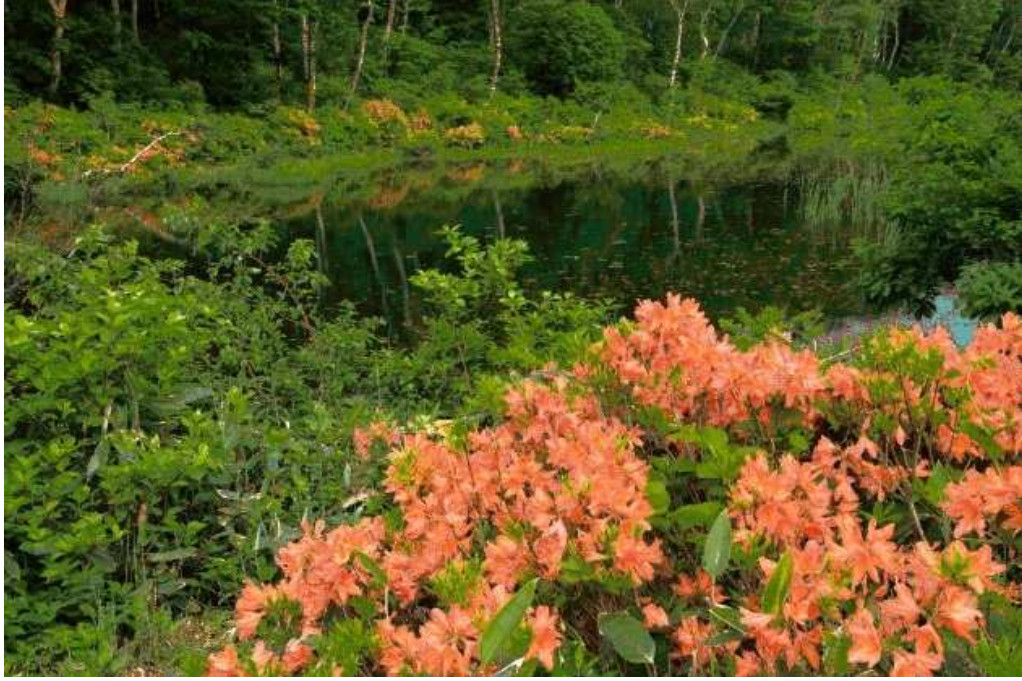}
}
\end{figure*}

\section{Summary}
We introduced Riemann normal coordinates as a tool to construct  mappings between the color spaces with metrics defined by the perception properties of different observers. These metrics were measured for color-weak and color-normal observers and 2D and 3D color-weak compensation were implemented. 
The semantic differential evaluation showed that both the 2D and the 3D compensations increased  correlation between impressions 
of the color-normal and the color-weak observer  and that the 3D compensation is more efficient than the 2D compensation.

\bibliographystyle{IEEEtran}
\bibliography{./tip}

\end{document}